\documentclass[10pt,twocolumn,letterpaper]{article}

\usepackage{cvpr}              
\usepackage{makecell}
\usepackage[table]{xcolor}
\usepackage{multirow}
\usepackage{stfloats}
\usepackage{bm}
\usepackage{pifont}
\usepackage{graphicx}
\usepackage{bbding}
\usepackage{tabularx}
\usepackage{booktabs}

\usepackage{adjustbox}

\usepackage{booktabs}   
\usepackage{multirow}   
\usepackage{adjustbox}  
\usepackage{amsmath}    
\usepackage{algorithm}
\usepackage{algpseudocode}

\definecolor{cvprblue}{rgb}{0.21,0.49,0.74}
\usepackage[pagebackref,breaklinks,colorlinks,allcolors=cvprblue]{hyperref}

\def\paperID{} 
\def\confName{CVPR}
\def\confYear{2026}

\title{Repurposing 3D Generative Model for Autoregressive Layout Generation}

\author{
    Haoran Feng\textsuperscript{1, 2}\footnotemark[1] \quad
    Yifan Niu\textsuperscript{1}\footnotemark[1] \quad
    Zehuan Huang\textsuperscript{1 \Envelope} \quad
    Yang-Tian Sun\textsuperscript{3} \\
    Yuxin Peng\textsuperscript{4} \quad
    Lu Sheng\textsuperscript{1 \Envelope} \\[0.3em]
    \textsuperscript{1}School of Software, Beihang University \quad
    \textsuperscript{2}Tsinghua University \\[0.2em]
    \textsuperscript{3}University of Hong Kong \quad
    \textsuperscript{4}Peking University \\\\[0.1em]
    Project page: \url{https://fenghora.github.io/LaviGen-Page/}
}

\begin{document}

\twocolumn[
    \maketitle
    \vspace{-1.5em}
    \begin{center}
\centering
\vspace{-0.05in}
\includegraphics[width=\textwidth]{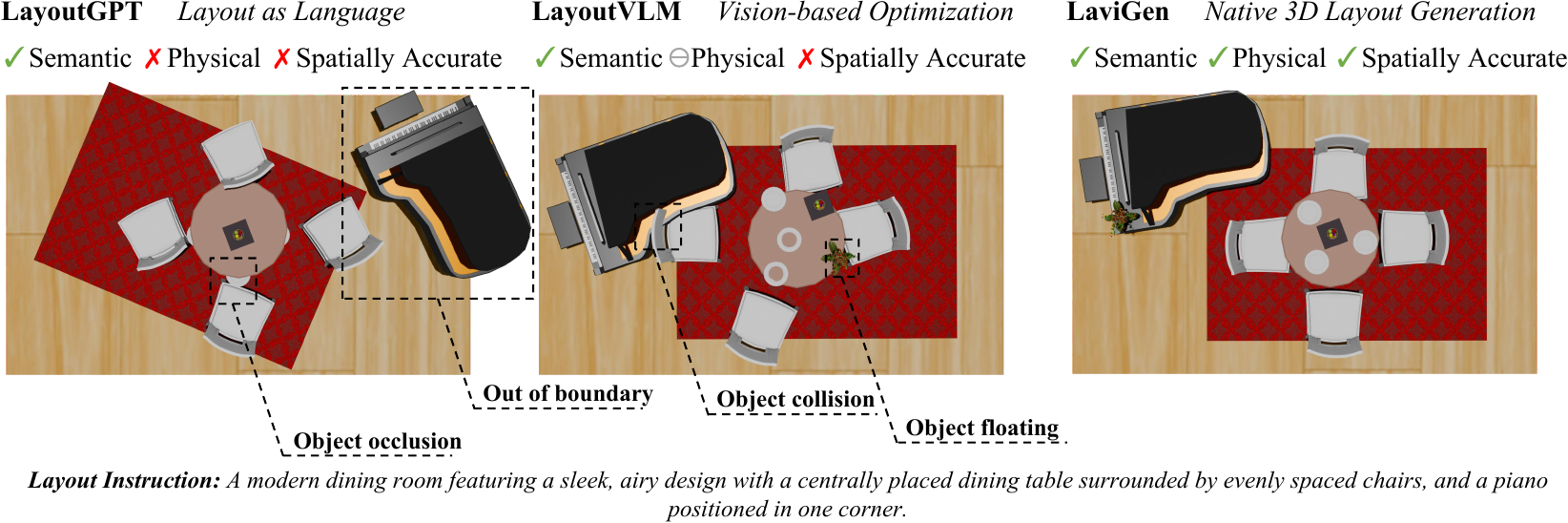}
\captionof{figure}{
\textit{LaviGen} generates layouts that are both physically plausible and semantically coherent from only 3D objects and instructions, whereas two baselines struggle.
Our framework achieves that by leveraging the 3D prior knowledge of generative models to perform generation directly in the native 3D space.
}
\label{fig:teaser}
\end{center}
    \bigbreak
]

\maketitle

\let\thefootnote\relax\footnotetext{
$^*$ Equal contribution \hspace{5pt}
\ding{41} Corresponding author
}

\begin{abstract}

We introduce \textit{LaviGen}, a framework that repurposes 3D generative models for 3D layout generation.
Unlike previous methods that infer object layouts from textual descriptions,
\textit{LaviGen} operates directly in the native 3D space, 
formulating layout generation as an autoregressive process that explicitly models geometric relations and physical constraints among objects, 
producing coherent and physically plausible 3D scenes.
To further enhance this process, we propose an adapted 3D diffusion model that integrates scene, object, and instruction information and employs a dual-guidance self-rollout distillation mechanism to improve efficiency and spatial accuracy.
Extensive experiments on the LayoutVLM benchmark show \textit{LaviGen} achieves superior 3D layout generation performance, with 19\% higher physical plausibility than the state of the art and 65\% faster computation.
Our code is publicly available at \href{https://github.com/fenghora/LaviGen}{https://github.com/fenghora/LaviGen}.

\end{abstract}    
\section{Introduction}
\label{sec:intro}

Generating coherent 3D scene layouts ~\cite{atiss, lin2024instructscene, yang2024physcene, scenegen, ling2024scenethesis, Gu2025ArtiSceneLA, SpatialLM, Yang2025LLMdrivenIS, idesign, holodeck,midi, tang2025geometrictexturalconsistency3d, huang2025literealitygraphicsready3dscene, liu2025agentic, yu2025metascenes} is essential for creating realistic and interactive VR/AR environments. 
It aims to arrange objects in semantically consistent and physics-compliant configurations, such as placing chairs around a table instead of against walls.
%
A central challenge, therefore, lies in effectively encoding the geometric distributions describing spatial relationships and semantic dependencies among objects in 3D space.

Early approaches~\cite{atiss} rely on limited 3D scene data with insufficient knowledge about real spatial relationships, and thus lead to physically implausible scene layouts.
Recent methods~\cite{layoutgpt,holodeck,idesign, Yang2025LLMdrivenIS,SpatialLM, ling2024scenethesis} such as LayoutGPT~\cite{layoutgpt} treat layout as language in a structured, JSON-like format, which can be generated by large language models (LLMs)~\cite{gpt4o,dubey2024llama,comanici2025gemini,achiam2023gpt}.
While rich language prior from the LLMs have been employed, 
the absence of physical modeling often leads to spatially inconsistent layouts, resulting in object collisions, inter-penetrations, or floating.
To address this limitation, LayoutVLM~\cite{layoutvlm} leverages visual signals to indirectly supervise layout generation, enhancing the visual plausibility of the resulting scenes.
However, image-level supervision is computationally costly and lacks a fundamental understanding of 3D spatial structures.


Inspired by the observation that scene layout is a special type of geometric distribution, we pose the question: 
\textit{Can layout generation be learned directly from geometric distributions of 3D scenes?}
Recent progress in 3D generative modeling~\cite{trellis,midi,scenegen} has made this feasible, offering powerful 3D priors that encode spatial relationships and geometric distributions. 
A key challenge, consequently, is how to harness these 3D priors, inherently providing spatial coherence, to enable layout generation, completion, and editing, tasks that are beyond the reach of previous text-based methods.
In this context, we repurpose 3D generative models for autoregressive layout generation: 
leveraging their built-in geometric priors about common spatial layouts,
the model places objects sequentially to produce updated scene states that inherently satisfy physically plausible spatial arrangements.
Compared to monolithic scene generation, where injecting all object conditions at once can destabilize the generation process, the autoregressive paradigm provides greater controllability and inherently supports object addition and removal.
%

However, building a 3D generative model into an autoregressive layout generation system is non-trivial. 
On the one hand, the generative model must simultaneously perceive and learn to align with both the global space of the scene geometry and the object's own canonical space. 
On the other hand, autoregressive generation inherently suffers from exposure bias~\cite{huang2025self}; 
when generating long sequences, this approach inevitably introduces accumulated spatial errors.


To address these challenges, we propose \textit{LaviGen}, a framework that repurposes 3D generative models for autoregressive layout generation.
%
%
Given an initial scene and an object, \textit{LaviGen} encodes them and integrates semantic information, then generates an updated scene by placing the object in a semantically consistent manner within the native 3D space.
In addition, a post-training strategy is proposed to mitigate exposure bias.
\textit{LaviGen} introduces a dual-guidance self-rollout distillation strategy that combines scene-level holistic guidance with step-wise scene-object alignment supervision, 
mitigating error accumulation in long-sequence generation and improving spatial coherence.

\begin{figure}[t]
    \centering
    \includegraphics[width=\linewidth]{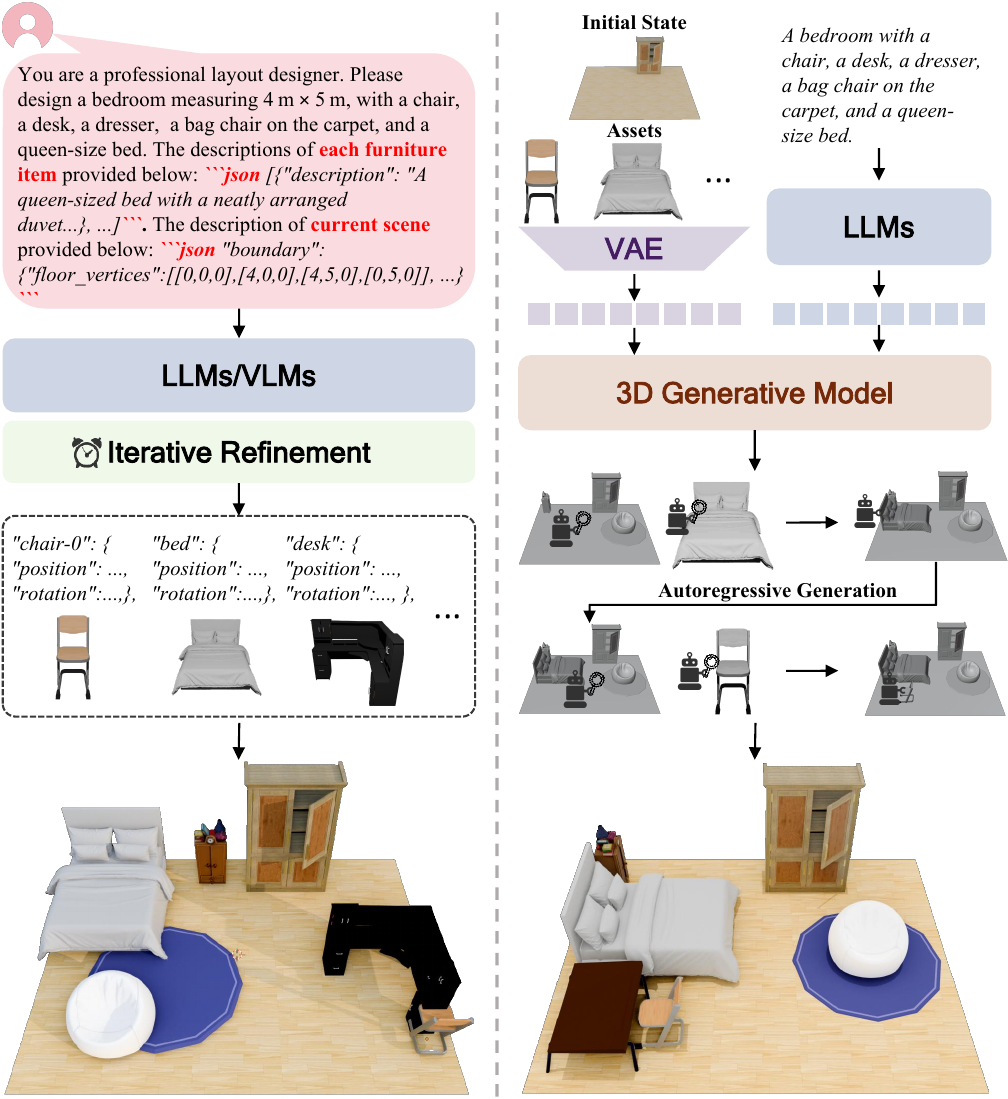}
    \caption{Our layout generation pipeline \emph{versus} existing methods that treat layouts as language or rely on vision-based optimization.}
    \label{fig:method_comparision}
    \vspace{-0.05in}
\end{figure}

As shown in~\cref{fig:method_comparision}, 
\emph{LaviGen} leverages the geometric priors of 3D scenes to autoregressively place objects in sequence, achieving more geometrically consistent and plausible layout reasoning while avoiding time-consuming iterative refinement~\cite{layoutvlm}.
Extensive experiments on the benchmark proposed by LayoutVLM~\cite{layoutvlm} demonstrate that \textit{LaviGen} outperforms existing layout generation approaches, 
achieving 19\% higher physical plausibility than the state of the art and reducing computational time by roughly 65\%.
%
It also supports a broader range of applications, such as layout completion and layout editing, which are difficult to achieve without operating in native 3D space.
The main contributions of this work are summarized as follows:

\begin{itemize}
    \item We propose \textit{LaviGen}, a framework that repurposes a 3D generative model for autoregressive layout generation, enabling layout synthesis directly in the native 3D space.
    \item Adapted 3D diffusion model and dual-guidance self-rollout distillation to capture environment-object contextual relationships and mitigate exposure bias.
    \item Experiments show that \textit{LaviGen} achieves superior physical plausibility and generalization in layout synthesis and extends naturally to layout completion and layout editing.
\end{itemize}

\section{Related Work}
\label{sec:related_work}

\vspace{+1.5mm}
\noindent\textbf{3D Layout Generation.}
Generating coherent 3D layouts is a long-standing challenge~\cite{atiss, lin2024instructscene, yang2024physcene, scenegen, ling2024scenethesis, Gu2025ArtiSceneLA, SpatialLM, Yang2025LLMdrivenIS, idesign, holodeck}.
Early learning-based methods, represented by ATISS~\cite{atiss}, employed autoregressive transformers to directly regress object placements. 
This direct coordinate prediction, however, often neglected geometric semantics, leading to spatial inconsistencies.
Subsequent approaches turned to foundation models, initiated by methods that treat layout as a language task~\cite{layoutgpt,holodeck,idesign, Yang2025LLMdrivenIS,SpatialLM, ling2024scenethesis}. 
These models leverage LLMs to output structured textual plans, excelling at semantic coherence. 
A noted limitation, however, is the difficulty in capturing explicit physical constraints, resulting in object collisions or floating artifacts. 
To address these physical inconsistencies, LayoutVLM~\cite{layoutvlm} introduced 2D visual supervision, using rendered images and differentiable optimization to refine poses. 
While this improves plausibility, this 2D supervision is not fully holistic for complex 3D interactions and introduces computationally expensive optimization. 
Both paradigms operate in non-native representations. In contrast, LaviGen formulates the task as a native 3D autoregressive process, operating directly in 3D space to explicitly model geometric relations and physical constraints from the ground up.

\begin{figure*}[t]
    \centering
    \includegraphics[width=\linewidth]{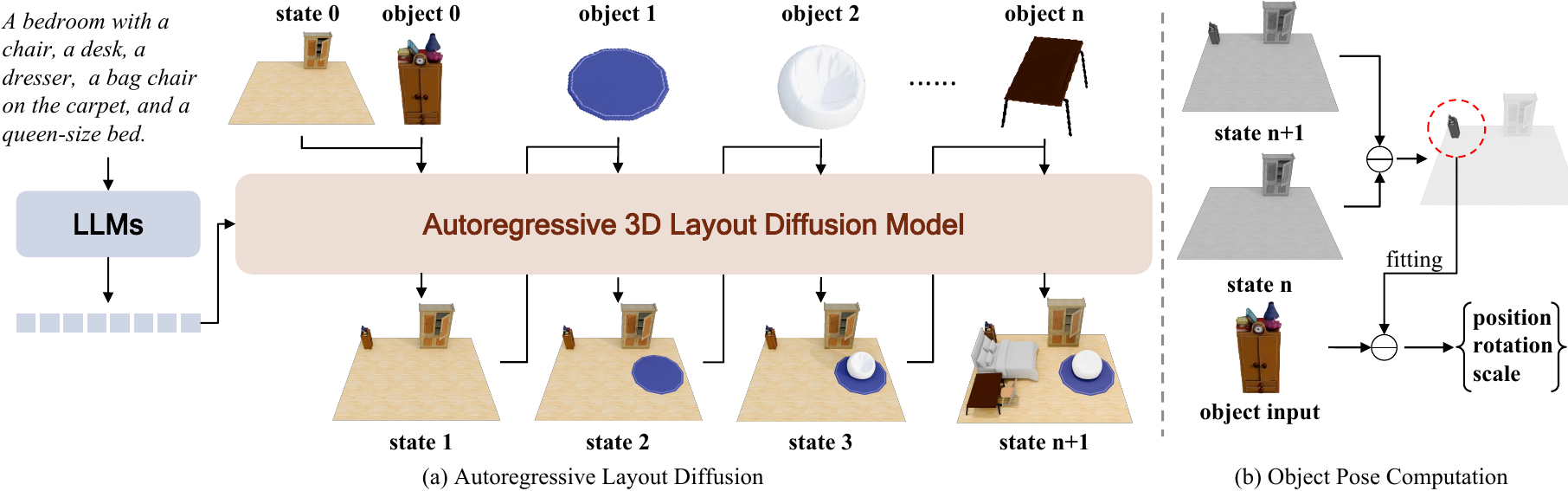}
    \caption{ \textbf{Overview of the \textit{LaviGen} framework for autoregressive 3D layout generation.} (a) LaviGen formulates layout generation as an autoregressive process. Specifically, conditioned on LLM-encoded instructions, it takes the current scene state $S_i$ and object $O_i$ to generate the updated state $S_{i+1}$. (b) The high-fidelity scene is obtained by computing the spatial difference between $S_{i+1}$ and $S_i$ to locate the newly generated region, and fitting the object $O_i$ to derive its spatial parameters.}
    \label{fig:pipeline}
    \vspace{-0.05in}
\end{figure*}

\vspace{+1.5mm}
\noindent\textbf{3D Generative Models.}
Recent development in diffusion models~\cite{ho2020ddpm,song2020ddim} and large-scale 3D datasets
~\cite{deitke2023objaverse,deitke2024objaversexl} has greatly advanced the field of 3D generation
~\cite{liu2024one2345,liu2023syncdreamer,long2024wonder3d,hong2023lrm,tang2025lgm,huang2024epidiff,zhang2024clay,wu2024unique3d,li2024craftsman,wen2024ouroboros3d,xu2024instantmesh,voleti2025sv3d,wang2024crm,liu2024one2345++,wu2024direct3d,zhao2024michelangelo,roessle2024l3dg,wu2024blockfusion,meng2024lt3sd,liu2024part123,dong2025tela,chen2024meshxlneuralcoordinatefield,chen2024meshanythingartistcreatedmeshgeneration,wang2024llamameshunifying3dmesh,hao2024meshtronhighfidelityartistlike3d,he2024neurallightrigunlockingaccurate,gao2025meshartgeneratingarticulatedmeshes,zhao2025deepmeshautoregressiveartistmeshcreation,wei2025octgptoctreebasedmultiscaleautoregressive,li2025step1x3dhighfidelitycontrollablegeneration,ye2025shapellmomninativemultimodalllm,nash2020polygen,ibing2023octree,li2024pasta,lu2025uni3dar}.
A series of research
~\cite{liu2023syncdreamer,long2024wonder3d,tang2025lgm,wen2024ouroboros3d,xu2024instantmesh,wang2024crm,voleti2025sv3d,huang2024mvadapter,qu2025deocc1to33ddeocclusionsingle,huang2025stereogsmultiviewstereovision} 
generates multi-view images and then reconstructs 3D assets, but two-stage bias often degrades geometric and textural fidelity.
A growing body of work
~\cite{zhang2024clay,li2024craftsman,wu2024direct3d,zhao2024michelangelo,li2025triposg,chen2025ultra3defficienthighfidelity3d,dong2025morecontextuallatents3d,zhao2025assemblerscalable3dassembly,tang2025efficientpartlevel3dobject,lin2025partcrafterstructured3dmesh,wu2025dipodualstateimagescontrolled,wu2025direct3ds2gigascale3dgeneration,li2025triposghighfidelity3dshape,trellis,li2025craftsman3dhighfidelitymeshgeneration} 
has explored native 3D diffusion architectures, typically combining a variational autoencoder~\cite{vae} for latent encoding with a diffusion transformer (DiT)~\cite{dit} for structured denoising in 3D space. 
Exhibiting high 3D fidelity and structural consistency, these models learn rich spatial relationships from large-scale 3D data, 
offering strong geometric priors that underpin our accurate and physically consistent layout generation.

\vspace{+1.5mm}
\noindent\textbf{Autoregressive Diffusion and Distillation.}
The sequential placement of objects naturally frames layout generation as a long-sequence autoregressive generation problem.
However, conventional diffusion models with bidirectional attention perform poorly on such autoregressive tasks~\cite{atiss, ge2022long, hong2022cogvideo, kondratyuk2024videopoet}.
Concurrently, the autoregressive generation process inherently suffers from exposure bias~\cite{huang2025self}, leading to accumulated errors~\cite{chen2024diffusion, chen2025skyreels, gu2025long, magi1, yin2025causvid}.
To alleviate this, Diffusion Forcing~\cite{gao2025ca2, hu2024acdit, jin2024pyramidal} trains models to denoise tokens conditioned on ground-truth context with independently sampled noise levels.
Self Forcing~\cite{huang2025self, yang2025longlive, huang2025memory, liu2025rollingforcingautoregressivelong, cui2025self} further improves stability by performing autoregressive rollouts during training, conditioning on the model’s own outputs.
We adopt a similar distillation-based autoregressive mechanism to enhance both efficiency and stability.

\section{Methodology}

As illustrated in~\cref{fig:pipeline}, \textit{LaviGen} is a unified framework that repurposes a pretrained 3D generative model for language-conditioned 3D layout synthesis.
Leveraging structured 3D priors, our framework ensures spatial coherence and physical plausibility, substantially reducing object collisions and boundary violations.
We begin by revisiting structured 3D latent models~\cite{trellis} in~\cref{sec:preliminary}, 
which form the geometric backbone of our approach.
We then detail the overall generative pipeline in~\cref{sec:llm_guided_3d_generative_model}, 
the autoregressive layout generation mechanism in~\cref{sec:auto_layout_gen}, 
and the dual-guidance self-rollout scheme at post-training in~\cref{sec:self_cond_distill}, 
which jointly enable efficient, accurate, and editable 3D layout synthesis.

\subsection{Preliminary}
\label{sec:preliminary}

\paragraph{Structured 3D generative models.}

The 3D prior module in \textit{LaviGen} draws inspiration from structured 3D latent diffusion models TRELLIS~\cite{trellis}, 
which typically generate 3D assets through a two-stage denoising process that first reconstructs coarse spatial structures and then refines fine-grained geometry and appearance. 
\textit{LaviGen} retains only the structure-level generation stage, predicting sparse voxel occupancies to model the spatial organization of objects and capture physically and semantically plausible spatial relationships.
Each 3D asset is represented by a set of voxel-indexed local latent codes
\begin{equation}
\mathcal{Z} = \{ z_{p} \mid p \in \mathcal{P} \},
\end{equation}
where \(\mathcal{P}\) denotes the set of active voxel positions near the object surface,
and each \(z_p \in \mathbb{R}^d\) is the local latent attached to voxel \(p\).
This structured representation allows for accurate modeling of 3D space.
For generation, TRELLIS adopts Flow Matching models, which add noise $\epsilon$ to clean data samples $x_0$ through $x(t) = (1 - t)x_0 + t\epsilon$ over time step $t$.
The reverse dynamics are expressed as a time-dependent vector field $v(x, t) = \nabla_t x$, 
which is learned via a neural approximation $v_\theta$ by minimizing the flow matching loss:
\begin{equation}
\mathcal{L} = \mathbb{E}_{t, x_0, \epsilon}  \left\| v_{\theta}(x, t) - (\epsilon - x_0) \right\|_2^2 .
\end{equation}

\subsection{LaviGen for 3D Layout Generation}
\label{sec:llm_guided_3d_generative_model}

In this work, we introduce \textit{LaviGen}, a  3D generative model for autoregressive layout generation that fundamentally differs from prior approaches~\cite{layoutgpt,atiss,layoutvlm,holodeck,idesign} predicting the spatial coordinates of objects from textual descriptions.
Our framework directly models the spatial configuration of objects in the native 3D space, 
enabling the generation of layouts that are both physically plausible and semantically coherent with textual descriptions.

As illustrated in~\cref{fig:pipeline}, 
given the current state \(S_i\), a target object \(O_i\), and the corresponding layout instruction, 
\textit{LaviGen} generates a physically and semantically plausible updated layout \(S_{i+1}\) directly within the 3D space, 
which then serves as the initial state for subsequent generation steps. 
Specifically, the layout instruction is encoded into a conditioning vector \(c\) for subsequent generation steps. 
Subsequently, during each generation step, the current state \(S_i\) and the target object \(O_i\) are encoded and concatenated with a noise latent, 
which are then fed into an autoregressive 3D layout diffusion model. 
The model then performs denoising conditioned on \(c\) through cross-attention, producing the updated scene state \(S_{i+1}\). 
This new state is then combined with the next object \(O_{i+1}\), and the process is repeated in an autoregressive manner to synthesize a coherent 3D layout sequence.
Finally, to reconstruct a high-fidelity 3D scene, we extract surface points from the high-resolution voxel occupancy decoded by the VAE.
We then downsample the original furniture meshes and register them to the extracted surface points via Iterative Closest Point, estimating optimal rotation, scale, and translation parameters through least-squares fitting.
The aligned objects are then placed into the generated layout to obtain the final scene.

\subsection{Autoregressive 3D Layout Diffusion}
\label{sec:auto_layout_gen}

The autoregressive layout generation enables the model to reconstruct the given scene $S$ and integrate a new object $O$ into it, producing spatially coherent and physically plausible layouts.
To this end, we design two key components: an \textbf{adapted 3D diffusion model} and an \textbf{identity-aware embedding module}.
These designs collectively enable the model to understand the current scene context and the 3D geometric properties of objects, generating physically plausible and semantically coherent layouts in native 3D space.

\vspace{+2mm}
\noindent\textbf{Architecture Adaptation.}
To enable object placement, we adapt the original 3D diffusion model by integrating scene, object, and noisy latents into a unified latent space to capture structured geometric relationships.
Specifically, during training, the scene $S$ and object $O$ are first encoded into latent representations $s$ and $o$, respectively, 
where $s, o \in \mathbb{R}^{N \times d}$, with $N = H \times W \times L$ representing the total number of latent voxels in a grid of height $H$, width $W$, and length $L$, while $d$ denotes the feature dimension of each voxel.
The latent representation of the target scene $x_0$ is perturbed with a randomly sampled noise $\epsilon \in \mathbb{R}^{N \times d}$, matching the shape of $s$ and $o$, and then concatenated with them and processed together with the textual embedding $c$ to guide denoising through semantic conditioning.

The overall training objective is
\begin{equation}
\mathcal{L}= \mathbb{E}_{t, x_0,s,o,c,\epsilon}  \left\| v_{\theta}(x, s,o,c, t) - (\epsilon - x_0) \right\|_2^2 .
\end{equation}
%

\vspace{+2mm}
\noindent\textbf{Identity-aware Positional Embedding.}
Although the adapted 3D diffusion model enables interactions among scene, object, and latent tokens, distinguishing the current scene state from newly added objects remains challenging. 
To mitigate this, an identity-aware embedding is introduced to explicitly encode the source identity of each token.
Concretely, we assign identical positional encodings to the noisy latent $x$ and the state $s$, reflecting their shared spatial coordinates, 
while the object $o$ receives a distinct encoding to preserve its individual geometric semantics.
%
This is implemented by extending the standard Rotary Position Embedding (RoPE)~\cite{rope} with an additional identity flag $f$ indicating the source of each token.
After concatenating the input $[x, s, o]$, each token is associated with a voxel at position $(f, h, w, l)$, where $f=0$ for the noisy latent $x$ and state $s$, and $f=1$ for the object  $o$. 
The spatial coordinates $(h, w, l)$ denote the voxel’s position within its respective latent grid. 
The complex-valued positional frequencies are computed as
\begin{equation}
\Phi(f,h,w,l) = [\,\phi_f(f); \, \phi_h(h); \, \phi_w(w); \, \phi_l(l)\,],
\end{equation}
where $\phi_f(f)$ encodes the latent source identity and $\phi_h, \phi_w, \phi_l$ follow the standard RoPE for spatial positions.
By embedding identity information in this manner, the model distinguishes different latent streams while preserving spatial alignment, thus enabling precise semantic disentanglement and geometry-consistent reasoning.

\begin{figure}[t]
    \centering
    \includegraphics[width=\linewidth]{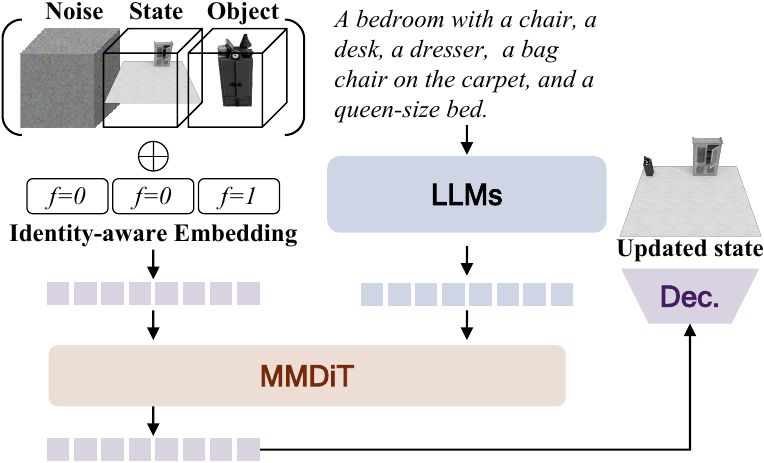}
    \caption{The overview of the \textbf{adapted 3D diffusion model}. 
    The encoded scene state and object are concatenated with the noisy latent and, together with the identity-aware embedding, fed into the multimodal diffusion transformer for noise prediction. The denoised output is then decoded to produce the updated scene state.
    }
\label{fig:autoregressive_generation}
\end{figure}

\subsection{Post-Training via Dual-Guidance Self-Rollout}
\label{sec:self_cond_distill}

While the autoregressive paradigm enables progressive scene composition,
it suffers from exposure bias:
the model is trained on ground-truth context but must condition on its own imperfect outputs at inference, causing errors to accumulate as collisions and implausible placements.
We mitigate this via self-rollout post-training, where the student autoregressively rolls out layouts from its own predictions during training, supervised by a holistic teacher for scene-level quality and a step-wise teacher for per-object accuracy.

\vspace{+2mm}
\noindent\textbf{Self-Rollout Mechanism.}
Inspired by Self-Forcing~\cite{huang2025self}, we introduce a self-rollout distillation framework.
In contrast to Teacher Forcing, which conditions on ground-truth context $S_{i-1}$, our method performs an autoregressive rollout during training, where the student $G_\theta$ conditions on its own generated layout $S_{i-1}^\theta$:
\begin{align}
\text{Teacher Forcing:} \quad & S_i^{\theta} = G_\theta(S_{i-1},\, O_i,\, c), \label{eq:teacher_forcing} \\
\text{Self-Rollout:} \quad & S_i^\theta = G_\theta(S_{i-1}^\theta,\, O_i,\, c), \label{eq:self_rollout}
\end{align}
where $i = 1, \ldots, n$ and $S_0^\theta{=}S_0$.
By replacing ground-truth conditioning with self-generated context, we force the model to encounter and learn to recover from its own errors, effectively bridging the train–test distribution gap and reducing error accumulation.
We instantiate $\mathcal{L}_{DM}$ as distribution matching distillation~\cite{yin2024improved}, minimizing the reverse KL divergence via score distillation with a learned critic model.
A key distinction from prior self-rollout methods in video generation~\cite{huang2025self} is that our autoregressive states are \emph{cumulative}, as each $S_i$ implicitly encodes all previously placed objects, unlike video frames which are independent rendering targets.
Consequently, per-frame supervision strategies used in video are insufficient here—errors in early placements propagate into all subsequent states, demanding supervision that addresses both the global scene quality and individual object placements.
This motivates our dual-guidance design below. We provide pseudocode in the supplementary material.


\vspace{+2mm}
\noindent\textbf{Holistic Guidance.}
Given this cumulative structure, a natural first approach is to supervise only the final scene $S_n^{\theta}$.
We use the bidirectional base model as a global planner $p_{\mathcal{T}_S}$, conditioned on text $c$, to provide holistic supervision over $S_n^{\theta}$ generated from $\mathcal{C}=(S_0,\{O_i\}_{i=1}^n,c)$:
\begin{equation}
\mathcal{L}_{holistic} = \mathcal{L}_{DM} \left( p_{\theta}(S_n | \mathcal{C}) \,||\, p_{\mathcal{T}_S}(S_n | c) \right)
\end{equation}
However, this alone proved insufficient, since supervision at only the terminal state provides no intermediate correction, and the teacher $p_{\mathcal{T}_S}$, not conditioned on object $O_i$, offers scene-level but not object-placement guidance.

\vspace{+2mm}
\noindent\textbf{Step-Wise Guidance.}
To provide dense, object-aware supervision at every step, we use the causal autoregressive model from \cref{sec:auto_layout_gen} as a per-step teacher $p_{\mathcal{T}_P}$.
Conditioned on $O_i$, it provides corrective signals at each step $i$ based on the student’s imperfect context. Let $\mathcal{C}_i=(S_{i-1}^\theta, O_i, c)$:
\begin{equation}
\mathcal{L}_{step} = \sum_{i=1}^{N} \mathcal{L}_{DM} \left( p_{\theta}(S_i | \mathcal{C}_i) \,||\, p_{\mathcal{T}_P}(S_i | \mathcal{C}_i) \right)
\end{equation}
Our final dual-guidance objective combines both terms with equal weights:
\begin{equation}
\mathcal{L}_{dual} = \mathcal{L}_{holistic} + \mathcal{L}_{step}.
\end{equation}
Concretely, $G_\theta$ is updated to align with the teacher score $s_{\mathcal{T}}$ via:
\begin{equation}
\nabla_{\theta}\mathcal{L}_{dual} \approx \mathbb{E}_{x_t, t} [ (s_{\mathcal{T}}(x_t, t) - s_{\psi}(x_t, t)) \nabla_{\theta} x_0 ],
\end{equation}
where $s_{\psi}$ is the critic score approximating the student distribution.
By training on self-rolled-out sequences, the student is directly exposed to its own error distribution; the holistic and step-wise teachers then provide complementary corrective signals at the scene and object level, respectively.
Full mathematical derivations and pseudocode are provided in the supplementary material.
    \begin{figure*}[t]
    \centering
    \includegraphics[width=\linewidth]{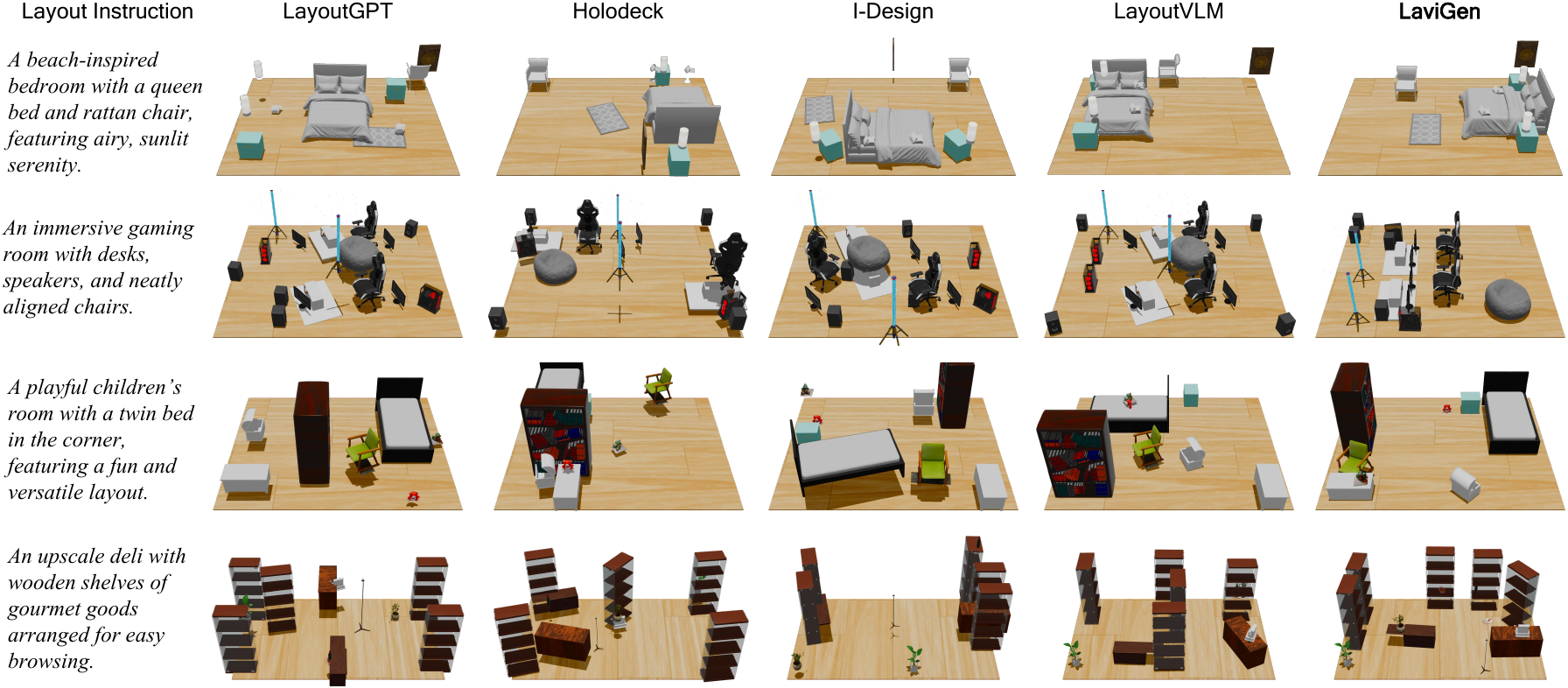}
    \caption{\textbf{Qualitative comparison of text-to-3D-layout generation.} \textit{LaviGen} produces physically plausible and spatially coherent layouts aligned with text prompts, effectively avoiding common failure cases such as object collisions (e.g., “gaming room”) and floating artifacts (e.g., “children’s room”, “deli”) observed in baselines.
    }
    \label{fig:comparision}
    
\end{figure*}

\section{Experiments}

\subsection{Setting}

\vspace{+2mm}
\noindent\textbf{Implementation Details.}
We implemented \textit{LaviGen} with an architecture inspired by Trellis~\cite{trellis}, reusing its structured variational autoencoder module, 
and trained the model from scratch using a three-stage training process.
First, we replace the original text encoder with Qwen2.5-VL-7B-Instruct~\cite{qwen2.5vl}, keeping its parameters frozen, and train the remaining model as the base bidirectional 3D generative model.
Next, we adopted an autoregressive generation paradigm to train the teacher model, which serves as the foundation for subsequent distillation and efficient inference.
Finally, we conduct dual-guidance self-rollout distillation: the teacher is distilled into a few-step student, which is then trained with our hybrid objective using the bidirectional model as the holistic teacher and the causal model as the step-wise teacher.
For autoregressive ordering, the Qwen-VL encoder derives a semantic object sequence from the input instruction during training, while inference supports either this learned order or user-defined sequences such as bottom-up placement.
Our model employs a 3B-parameter DiT and converges stably without delicate hyperparameter tuning, with 20 epochs for base training, 20k steps for teacher fine-tuning, and 5k steps for distillation.

\vspace{+2mm}
\noindent\textbf{Datasets.}
For the first training stage, we use the same dataset as Trellis~\cite{trellis}, which contains approximately 500K high-quality 3D assets collected from four public datasets: 
Objaverse-XL~\cite{objaversexl}, ABO~\cite{abo}, 3D-FUTURE~\cite{3dfuture}, and HSSD~\cite{hssd}.
For each 3D model, we first render its corresponding images and then use GPT-4o~\cite{gpt4o} to generate semantically rich annotations.
In the second and third stages, we train our model on two large-scale scene datasets, 3D-FRONT ~\cite{3dfuture} and InternScenes~\cite{internscenes}, 
comprising about 15K high-quality layout scenes with well-structured spatial arrangements.
For fair comparison, we follow LayoutVLM~\cite{layoutvlm} to evaluate \textit{LaviGen} on the same benchmark.

\vspace{+2mm}
\noindent\textbf{Metrics.}
We follow LayoutVLM~\cite{layoutvlm} and evaluate 3D layouts in terms of physical plausibility, semantic alignment, and computational efficiency. 
Physical plausibility is quantified by the \textit{Collision-Free (CF)} and \textit{In-Boundary (IB)} scores, ensuring that objects are non-overlapping and remain within scene boundaries. 
Semantic alignment is assessed using \textit{Positional (Pos.)} and \textit{Rotational (Rot.) coherency}, which measure the consistency between the generated layout and the textual prompt. 
For layouts without ground truth, GPT-4o~\cite{gpt4o} provides semantic ratings from both top-down and side views. 
We also report the \textit{Physically-Grounded Semantic Alignment (PSA)} score introduced by LayoutVLM, which combines semantic relevance with physical feasibility. 
Finally, we include the average inference time (T) to evaluate computational efficiency.
As LayoutVLM~\cite{layoutvlm} exhibits notably higher latency with increasing object counts, we report results on layouts with 8–10 objects.
All metrics except \textit{T} are normalized to [0,100], where higher values indicate better performance; lower \textit{T (s)} indicates faster inference.

\begin{figure*}[t]
    \centering
    \includegraphics[width=\linewidth]{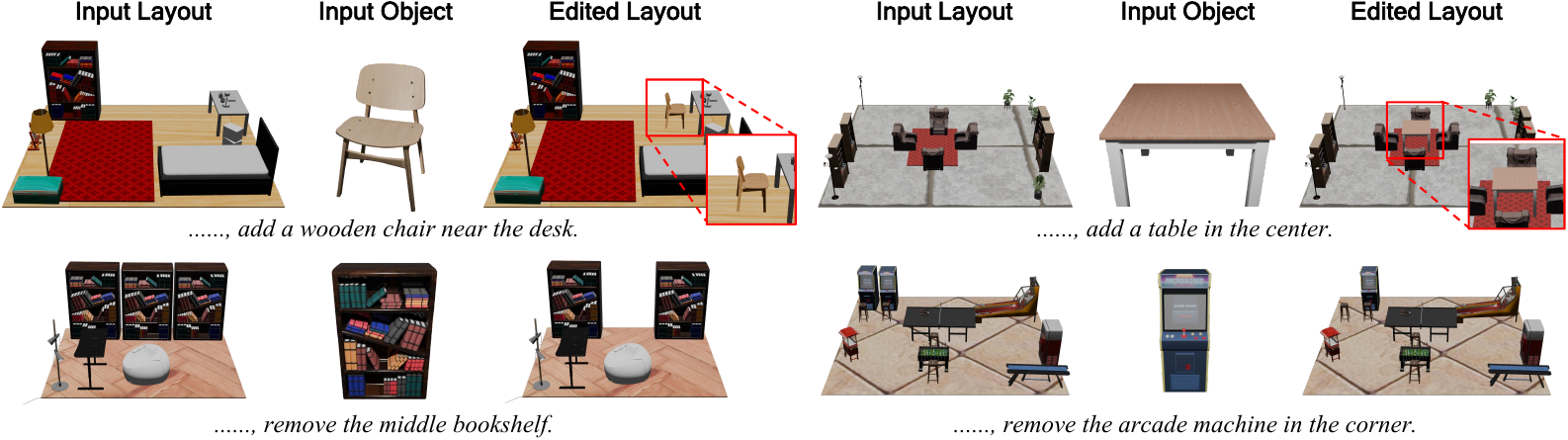}
    \caption{\textbf{Qualitative results for layout editing.} \textit{LaviGen's} unified framework supports context-aware modifications, including object insertion (top row) and removal (bottom row). By operating directly in 3D space, the model performs edits that are spatially coherent and semantically consistent with the surrounding context, and enables direct manipulation of 3D layouts, where previous methods struggle.}
    \label{fig:edit_results}
    
\end{figure*}

\subsection{Main Result}

\vspace{+2mm}
\noindent\textbf{Baselines.}
We compare our method with several state-of-the-art approaches for text-driven layout generation. 
LayoutGPT~\cite{layoutgpt} relies on a large language model to directly produce JSON-based layouts. 
Holodeck~\cite{holodeck} and I-Design~\cite{idesign} enhance this process with iterative optimization for layout refinement, 
while LayoutVLM~\cite{layoutvlm} further incorporates visual cues to improve generation quality.

\vspace{+2mm}
\noindent\textbf{Qualitative and Quantitative Comparison.}
As shown in~\cref{fig:comparision}, by operating directly in 3D space, \textit{LaviGen} accurately models inter-object relations and produces physically plausible arrangements with minimal collisions or floating artifacts, even in cluttered scenes like a gaming room where other baselines~\cite{layoutgpt,layoutvlm,holodeck,idesign} often fail.
Examining existing baselines, 
LayoutGPT~\cite{layoutgpt} generates layouts that are semantically coherent but frequently suffers from collisions, out-of-bound placements, and floating objects. 
Meanwhile, Holodeck~\cite{holodeck} struggles with arranging large objects, which negatively affects its performance.
Through iterative optimization, I-Design~\cite{idesign} partially mitigates these issues.
LayoutVLM~\cite{layoutvlm} takes a step further by leveraging rendered views to effectively address out-of-bound placements, though collisions and floating objects remain problematic.

Quantitative results in~\cref{tab:comparison} align with these observations.
\textit{LaviGen} achieves the best \textit{CF} and \textit{IB} scores, demonstrating remarkable physical plausibility.
LayoutGPT~\cite{layoutgpt} excels in semantic coherence but overlooks physical constraints, resulting in frequent collisions and boundary violations.
Holodeck~\cite{holodeck} performs poorly on geometric metrics, particularly the \textit{IB} score, as small parameter variations for large objects often cause severe placement errors.
I-Design~\cite{idesign} stands as the strongest text-only method, yet its iterative optimization incurs substantial computational cost.
LayoutVLM~\cite{layoutvlm}, benefiting from visual supervision, performs relatively well overall, though physical plausibility remains suboptimal and rendering introduces additional overhead.

These limitations stem from excessive information compression: representing objects solely with bounding boxes fails to capture fine-grained spatial interactions, and while LayoutVLM partially compensates with visual cues, it is constrained by the trade-off between rendered views and computational cost.
These findings underscore the advantage of generating layouts directly in native 3D space, where complete geometric information supports physically plausible and semantically coherent arrangements.

\begin{table}[t]
    \caption{\textbf{Main quantitative comparison and ablation study.} The top section compares \textit{LaviGen} against state-of-the-art baselines. The bottom section ablates the key components of our model, validating their progressive contributions to the final performance.}
    \small
    \label{tab:comparison}
    \centering
    \vspace{-1mm}
    \adjustbox{width={\linewidth}, keepaspectratio}{
    \begin{tabular}{l|cc|cc|cc} 
        \toprule
        \multirow{2}{*}{Methods} & \multicolumn{2}{c|}{Physical} & \multicolumn{2}{c|}{Semantic} & \multirow{2}{*}{PSA $\uparrow$} & \multirow{2}{*}{T (s) $\downarrow$} \\
        \cmidrule(lr){2-3} \cmidrule(lr){4-5} 
        & CF $\uparrow$ & IB $\uparrow$ & Pos. $\uparrow$ & Rot. $\uparrow$ & & \\
        \midrule
        LayoutGPT~\cite{layoutgpt} & 83.8 & 24.2 & \textbf{80.8} & \textbf{78.0} & 16.6 & \textbf{21.3}\\
        Holodeck~\cite{holodeck} & 77.8 & 8.1 & 62.8 & 55.6 & 5.6 & 58.2 \\
        I-Design~\cite{idesign} & 76.8 & 34.3 & 68.3 & 62.8 & 18.0 & 179.2 \\
        LayoutVLM~\cite{layoutvlm} & 81.8 & 94.9 & 77.5 & 73.2 & 58.8 & 75.5 \\
        \midrule
        base model & 75.6 & 64.8 & 45.1 & 44.7 & 16.7 & 145.7 \\
        + id-aware emb. & 89.1 & 96.8 & 68.8 & 66.5 & 71.4 & 144.1\\
        + $\mathcal{L}_{holistic}$. & 79.5 & 81.9 & 61.4 & 58.7 & 59.7 & 24.5 \\
        + $\mathcal{L}_{step}$. \textbf{(full)} & \textbf{97.3} & \textbf{98.6} & 76.9 & 77.1 & \textbf{78.8} & 24.3 \\
        \bottomrule
    \end{tabular}}
\end{table}

\begin{figure*}[t]
    \centering
    \includegraphics[width=\linewidth]{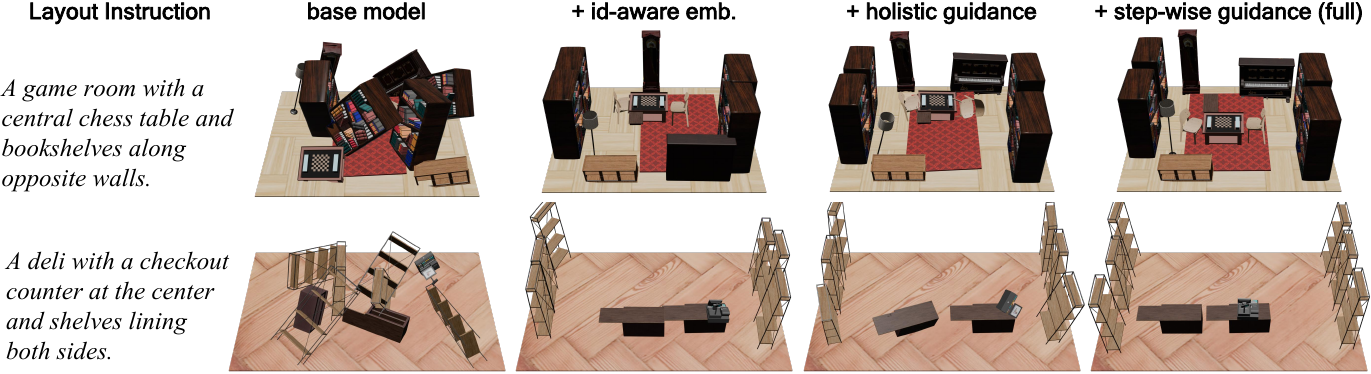}
    \caption{\textbf{Qualitative ablation study for \textit{LaviGen}.} We show the progressive improvement from the base model (left) to the full model (right). 
    The baseline produces cluttered layouts with severe collisions, while adding the identity-aware embedding yields a more plausible distribution but still suffers collisions from exposure bias. Distillation with holistic guidance yields inaccurate object fitting and severe inversion errors for small objects. In contrast, the full model generates physically plausible and semantically coherent layouts.}
    \label{fig:method_ablation}

\end{figure*}

\vspace{+2mm}
\noindent\textbf{User Study.}
To further evaluate our model, we conducted a user study with 43 participants, each answering 10 questions, yielding 430 responses. 
For each question, participants selected the best model based on physical plausibility, semantic consistency, and overall quality.
\cref{tab:user} shows that \textit{LaviGen} excels in physical plausibility and overall quality while maintaining comparable semantic consistency.

\subsection{Applications}

\vspace{+2mm}
\noindent\textbf{Layout Completion.}
Layout completion refers to the task of generating a complete layout for a partially specified scene, given layout instructions and 3D assets.
This scenario is common, as incomplete or unlabeled 3D layouts often result from limited annotations or missing metadata, making robust completion crucial for downstream tasks.
However, language-based approaches~\cite{layoutgpt,holodeck,idesign} struggle in these cases, as they rely on textual cues rather than 3D spatial understanding.
In contrast, \textit{LaviGen} successfully completes this task by operating directly in 3D space, placing objects into the current scene with physical plausibility and semantic coherence.
This capability makes it well-suited for applications such as robotic perception, AR/VR environment generation, and autonomous navigation, where comprehensive textual annotations are often unavailable.

\vspace{+2mm}
\noindent\textbf{Layout Editing.}
Layout editing is also a practically important capability, allowing users to interact directly with 3D scenes.
With a simple adjustment, \textit{LaviGen} can perform high-quality edits, as shown in~\cref{fig:edit_results}. 
To enable this capability, we modify the training paradigm by swapping the autoregressive targets, allowing the model to remove objects from a scene and regenerate them in a context-aware manner.
This formulation enables object removal, insertion, and replacement within a single framework, allowing the model to perform edits that are spatially coherent and semantically consistent with the surrounding context.
By operating directly in 3D space without relying on textual cues, \textit{LaviGen} further demonstrates strong generalization and practicality for real-world layout modification tasks.

\begin{table}[t]
    \caption{User study results for layout generation.}
    \small
    \label{tab:user}
    \centering
    \vspace{-1mm}
    \adjustbox{width={\linewidth},keepaspectratio}{
    \begin{tabular}{l|ccc}
        \toprule
        Methods & Phys. Plaus. $\uparrow$ 
                & Sem. Consist. $\uparrow$ 
                & Ovr. Qual. $\uparrow$  \\
        \midrule
        LayoutGPT~\cite{layoutgpt} & 16.0 & 38.8 & 7.9 \\
        LayoutVLM~\cite{layoutvlm} & 31.9 & 27.7 & 36.5 \\
        LaviGen & 52.1 & 33.5 & 55.6 \\
       \bottomrule
    \end{tabular}}
    \vspace{-2mm}
\end{table}

\subsection{Ablation Study}

To assess the contribution of each component,
we conduct ablation studies starting from the base 3D generative model and progressively adding the identity-aware embedding, holistic guidance, and step-wise guidance.
The qualitative and quantitative results are presented in \cref{fig:method_ablation} and \cref{tab:comparison}.
Initially, the baseline fails to correctly interpret object–scene relationships, producing cluttered and semantically inconsistent layouts with severe collisions.
Adding the identity-aware embedding improves layout coherence, but exposure bias still introduces extraneous points between objects, causing collisions; without distillation, inference also remains slow due to excessive steps.
When distilled with holistic guidance, 
the generation time is greatly reduced, 
but object-fitting accuracy suffers, 
particularly for small objects, leading to errors in rotation prediction and noticeable flipping artifacts.
Finally, with the introduction of step-wise guidance, i.e., the proposed \textit{LaviGen}, the model produces physically plausible and semantically consistent layouts.
These results demonstrate the advantage of generating layouts directly in native 3D space and validate the effectiveness of the proposed framework.

\section{Conclusion}

We introduced \textit{LaviGen}, a framework for autoregressive 3D layout generation that operates directly in native 3D space.
Unlike prior approaches that treat layout as language, LaviGen leverages the geometric priors encoded in 3D generative models, enabling physically plausible and semantically coherent layout generation.
We introduce an adapted autoregressive 3D layout diffusion model, fully modeling the spatial relationships between the current scene and input objects to generate an updated scene.
To mitigate exposure bias in long-sequence generation, we further propose dual-guidance self-rollout distillation, improving training stability and physical fidelity.
Experiments demonstrate that LaviGen achieves superior spatial accuracy and efficiency across baselines, 
highlighting how its 3D generative paradigm provides a principled foundation for geometry-aware, semantically controllable scene generation.


\clearpage
\setcounter{page}{1}
\setcounter{section}{5}  
\setcounter{figure}{7}  
\setcounter{table}{2}   
\maketitlesupplementary

\section{Implementation Details}

\subsection{Base 3D Generative Model}
\label{impl_base_model}

To equip our system with a strong and expressive 3D prior, we begin by training a base 3D generative model. 
Our design follows the state-of-the-art structured 3D generative framework TRELLIS~\cite{trellis}, particularly its structure-level generation stage, which predicts sparse voxel occupancies to capture the spatial organization of objects and to model physically and semantically plausible scene layouts.
Concretely, we reuse the structured variational autoencoder of TRELLIS as the backbone, providing a compact and expressive representation of 3D structures.
To further enhance the model’s ability to interpret complex semantic layouts, we adopt Qwen2.5-VL-7B-Instruct~\cite{qwen2.5vl} as the text encoder, ensuring rich cross-modal grounding. 
The design of our flow transformer builds upon the architecture of Qwen-Image~\cite{qwenimage} and integrates the Multimodal Diffusion Transformer (MMDiT)~\cite{mmdit}, 
thereby enabling unified modeling of text and 3D representations within a single Transformer framework.
Within each block of the MMDiT, we incorporate a novel positional encoding mechanism, Multimodal Scalable RoPE (MSRoPE), designed to provide consistent and scale-robust positional representations across both modalities. 
This formulation enables effective multimodal fusion while preserving stable positional semantics for text and scalable spatial modeling for 3D latent representations.

The detailed experimental settings are as follows. 
We represent the 3D scene using a $64^3$ voxel grid, which is used across all training stages as well as during inference.
For training, we adopt classifier-free guidance (CFG)~\cite{cfg} with a drop rate of 0.1 and use the AdamW optimizer~\cite{adamw} with a learning rate of $1\times10^{-4}$. 
The model is trained for 400K steps on 16 A100 GPUs (80GB) with a batch size of 16 per GPU. 
At inference time, the CFG strength is set to 3 and 50 sampling steps are used.

\subsection{Teacher Model}
\label{impl_teacher_model}
With a strong 3D prior established, the next stage focuses on applying it to layout generation. 
To preserve the spatial knowledge already acquired by the base model, we minimize modifications to the original architecture. 
Specifically, as illustrated in Sec.~3.3, the teacher model builds upon the base 3D generative model by jointly taking the current scene state and the target object as input. 
To enable the model to distinguish between the scene and objects while facilitating faster convergence, we further introduce an identity-aware positional embedding. 
Together, these designs allow the teacher model to achieve comprehensive modeling of the geometric relationships between the scene and objects.
The experimental setup is largely consistent with that of the first stage, with the main differences being a reduced training length of 100K steps and a learning rate of $5\times10^{-5}$.

\subsection{Post-Training via Dual-Guidance Self-Rollout}

To mitigate the exposure bias inherent in autoregressive generation, we employ a dual-guidance self-rollout strategy, as summarized in Algorithm~\ref{alg:dual_guidance}. This stage distills a pre-trained, few-step student generator following the methodology detailed in the main paper Sec.~3.4. Below, we specify the network components and hyperparameters used in this process.

The distillation framework comprises four distinct models.
The Student Model $G_{\theta}$ is an efficient, few-step 3D layout diffusion model, initialized via distillation from the Autoregressive Teacher Model trained in \cref{impl_teacher_model}.
The Holistic Teacher $p_{\mathcal{T}_S}$ provides the final supervision signal $\mathcal{L}_{holistic}$ and is implemented using the frozen Base 3D Generative Model described in \cref{impl_base_model}.
The Step-Wise Teacher $p_{\mathcal{T}_P}$ provides intermediate corrective signals $\mathcal{L}_{step}$ utilizing the frozen Autoregressive Teacher Model detailed in \cref{impl_teacher_model}.
Finally, to implement the Distribution Matching Distillation (DMD) loss, we employ a trainable Critic Model $f_\psi$. This critic is initialized with the same architecture and weights as the Base 3D Generative Model, i.e., the Holistic Teacher, and is trained to approximate the score function of the student's generated data distribution.

\paragraph{Training Hyperparameters.}
We perform post-training with a batch size of 1 given the sequential, memory-intensive nature of the self-rollout process.
The Student Model $G_\theta$ is optimized using AdamW with a learning rate of $2\times10^{-6}$, $(\beta_1, \beta_2)=(0.0, 0.999)$, and weight decay of $0.01$.
The Critic Network $f_\psi$ is optimized separately using AdamW with a learning rate of $5\times 10^{-7}$, $(\beta_1, \beta_2)=(0.0, 0.999)$, and weight decay of $0.01$.
To stabilize score estimation, we use a Generator/Critic update ratio of 1:5 (i.e., five critic updates per student update).
For teacher score computation $s_{\mathcal{T}}$, Classifier-Free Guidance (CFG) is applied with a scale of $3.0$.

\paragraph{Loss Function Formulation.}
Our dual-guidance objective $\mathcal{L}_{dual} = \mathcal{L}_{holistic} + \mathcal{L}_{step}$ is formulated using Distribution Matching Distillation~\cite{yin2024one}.
This objective minimizes the reverse Kullback-Leibler divergence by leveraging the score difference between the student (approximated by the critic $f_\psi$) and the teacher.
The gradient for the student $G_\theta$ is derived as follows:
\begin{equation}
    \nabla_\theta \mathcal{L}_{dual} \approx \mathbb{E}_{x_t, t} \left[ (s_{\mathcal{T}}(x_t, t) - s_{\psi}(x_t, t)) \nabla_{\theta} x_0 \right]
\end{equation}
where $x_0 = G_\theta(x_t, t, \mathcal{C}_i)$ denotes the clean layout predicted by the student.
Here, $s_{\mathcal{T}}$ represents the score function of the fixed teacher (either $p_{\mathcal{T}_S}$ for $\mathcal{L}_{holistic}$ or $p_{\mathcal{T}_P}$ for $\mathcal{L}_{step}$), and $s_{\psi}$ is the score estimated by the critic.
The critic is concurrently trained to approximate the student's score using a standard denoising objective.

\begin{algorithm}[ht]
    \caption{Dual-Guidance Self-Rollout Distillation}
    \label{alg:dual_guidance}
    \begin{algorithmic}[1]
        \Require Denoise timesteps $\{t_1, \ldots, t_T\}$, number of objects $N$
        \Require Student generator $G_\theta$, step-wise teacher $p_{\mathcal{T}_P}$, holistic teacher $p_{\mathcal{T}_S}$
        \Require Initial state $S_0$, object sequence $\{O_i\}_{i=1}^N$, text prompt $c$
        \Loop
            \State $S_\text{ctx} \leftarrow S_0$, $S_{\text{outputs}} \leftarrow []$
            \State Sample $s \sim \text{Uniform}(1, \ldots, T)$
            \For{$i = 1, \ldots, N$}
                \State $\mathcal{C}_i \leftarrow (S_\text{ctx}, O_i, c)$
                \State Initialize $z_{t_T} \sim \mathcal{N}(0, I)$
                \For{$j = T, \ldots, s$}
                    \If{$j == s$}
                        \State Enable gradient computation
                        \State $\hat{S}_0 \leftarrow G_\theta(z_{t_j}; t_j, \mathcal{C}_i)$
                        \State $S_{\text{outputs}}.\text{append}(\hat{S}_0)$
                        \State $S_\text{ctx} \leftarrow \hat{S}_0.\text{detach}()$
                    \Else
                        \State Disable gradient computation
                        \State $\hat{S}_0 \leftarrow G_\theta(z_{t_j}; t_j, \mathcal{C}_i)$
                        \State Sample $\epsilon \sim \mathcal{N}(0, I)$
                        \State $z_{t_{j-1}} \leftarrow \Psi(\hat{S}_0, \epsilon, t_{j-1})$
                    \EndIf
                \EndFor
            \EndFor
            \Statex
            \State $\mathcal{L}_{step} \leftarrow 0$, $S_\text{ctx} \leftarrow S_0$
            \For{$i = 1, \ldots, N$}
                \State $\hat{S}_i \leftarrow S_{\text{outputs}}[i]$
                \State $\mathcal{C}_i \leftarrow (S_\text{ctx}, O_i, c)$
                \State $\mathcal{L}_{step} \leftarrow \mathcal{L}_{step} + \mathcal{L}_\text{DMD}(\hat{S}_i; p_{\mathcal{T}_P}, \mathcal{C}_i)$
                \State $S_\text{ctx} \leftarrow \hat{S}_i.\text{detach}()$
            \EndFor
            \State $\hat{S}_N \leftarrow S_{\text{outputs}}.\text{last}()$
            \State $\mathcal{L}_{holistic} \leftarrow \mathcal{L}_\text{DMD}(\hat{S}_N; p_{\mathcal{T}_S}, c)$
            \State $\mathcal{L}_{dual} \leftarrow \mathcal{L}_{step} + \mathcal{L}_{holistic}$
            \State Update $\theta$ via $\nabla_\theta \mathcal{L}_{dual}$
        \EndLoop
    \end{algorithmic}
\end{algorithm}

\section{Additional Results}

\subsection{Scalability to More Objects}
\begin{figure}[ht]
    \centering
    \includegraphics[width=\linewidth]{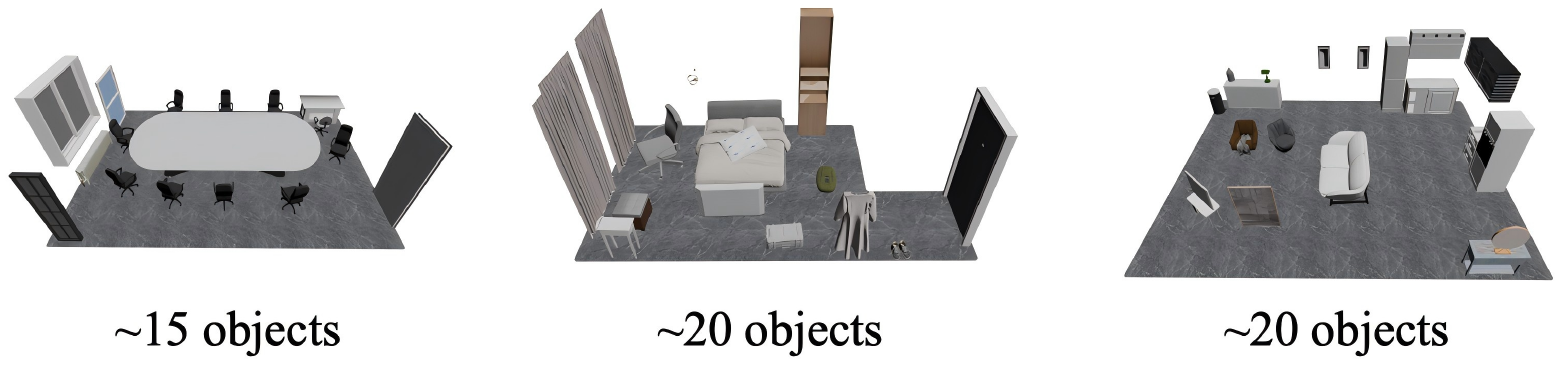}
    \vspace{-1.05in}
    \caption{Qualitative results for long-sequence generation with more than 20 objects.}
    \label{fig:multi_objects}
\end{figure}

We evaluate on 8--10 objects for fair comparison against existing baselines. 
Thanks to self-rollout distillation, LaviGen inherently supports a ``train short, test long'' paradigm and can handle scenes with more than 20 objects, as shown in~\cref{fig:multi_objects}.

\vspace{-0.9in}
\vspace{-0.2in}

\subsection{Generalizability Across Backbones}
\begin{figure}[ht]
    \centering
    \includegraphics[width=\linewidth]{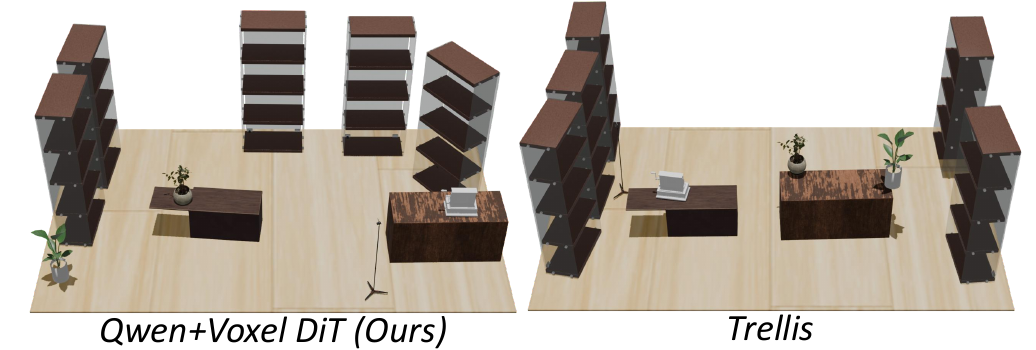}
    \caption{Generalization across different 3D generation backbones.}
    \label{fig:generalizability}
\end{figure}

Our framework is not tied to a specific 3D generative backbone. To validate this, we apply LaviGen to TRELLIS~\cite{trellis} using its original CLIP text encoder, without our additionally trained Qwen encoder. As shown in~\cref{fig:generalizability}, the model maintains high physical plausibility and semantic coherence, confirming that our recipe successfully transfers across different base architectures without relying on large-scale training infrastructure.

\vspace{-0.2in}
\subsection{Generation Diversity}
\begin{figure}[ht]
    \centering
    \includegraphics[width=\linewidth]{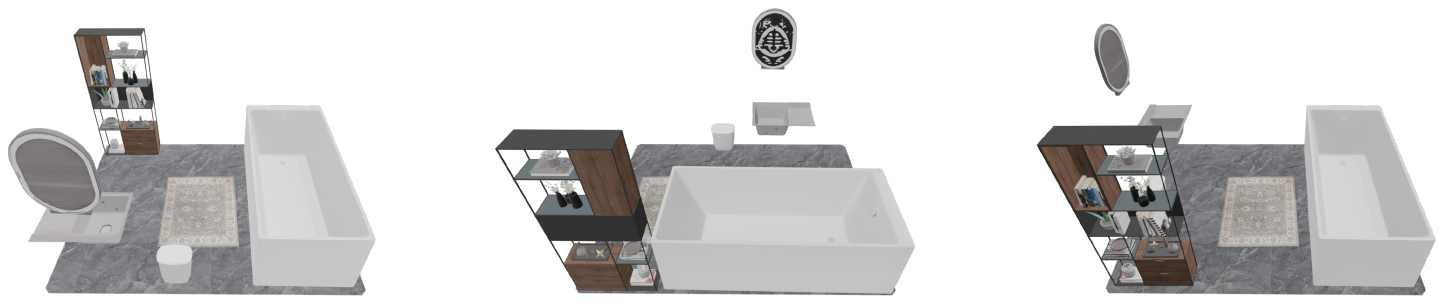}
    \vspace{-1.05in}
    \caption{Diverse layout results generated from the same input instruction.}
    \label{fig:diversity}
\end{figure}

LaviGen naturally supports diverse outputs via stochastic sampling. Given the same input instruction, the model generates varied yet plausible layouts, as illustrated in~\cref{fig:diversity}.

\vspace{-1in}

\vspace{-0.1in}
\section{Limitations and Future Work.}

Although \textit{LaviGen} gains strong geometric distribution modeling capability from operating in the native 3D space, several limitations remain.
First, due to constraints in model capacity and computational resources, we adopt a $64^3$ 3D grid resolution. 
While generally adequate for most objects, this resolution becomes insufficient for small instances, leading to mismatches in subsequent spatial coordinate computations. 
To address this issue, our future work will explore more efficient computation strategies for higher-resolution voxel grids and investigate denser 3D representations capable of supporting higher spatial resolutions and capturing finer spatial details.
Additionally, as shown in Tab.~1, the semantic consistency of the generated layouts remains suboptimal.
We attribute this primarily to the scarcity of high-quality annotations, particularly for layouts with complex spatial configurations or object arrangements. 
In future work, we will enhance our annotation pipeline to collect and process additional high-quality labeled data, and explore more advanced text-conditioning mechanisms to further improve the robustness and semantic reliability of \textit{LaviGen}.



\section*{Acknowledgments}
This work was supported by the grants from the National Natural Science Foundation of China (62132001, 62525201, 62432001) and Beijing Natural Science Foundation (L247006), and the Fundamental Research Funds for the Central Universities.
{
    \small
    \bibliographystyle{ieeenat_fullname}
    \bibliography{main}

@String(CVPR= {IEEE Conf. Comput. Vis. Pattern Recog.})

@String(ECCV= {Eur. Conf. Comput. Vis.})

@String(TOG= {ACM Trans. Graph.})

@String(ICLR = {Int. Conf. Learn. Represent.})

@String(AAAI = {AAAI})

@String(CVPR  = {CVPR})

@String(ECCV  = {ECCV})

@String(TOG   = {ACM TOG})

@String(ICLR  = {ICLR})

@inproceedings{trellis,
  title={Structured 3d latents for scalable and versatile 3d generation},
  author={Xiang, Jianfeng and Lv, Zelong and Xu, Sicheng and Deng, Yu and Wang, Ruicheng and Zhang, Bowen and Chen, Dong and Tong, Xin and Yang, Jiaolong},
  booktitle={Proceedings of the IEEE/CVF conference on computer vision and pattern recognition},
  year={2025}
}

@article{liu2024one2345,
  title={One-2-3-45: Any single image to 3d mesh in 45 seconds without per-shape optimization},
  author={Liu, Minghua and Xu, Chao and Jin, Haian and Chen, Linghao and Varma T, Mukund and Xu, Zexiang and Su, Hao},
  journal={Advances in Neural Information Processing Systems},
  volume={36},
  year={2024}
}

@article{liu2023syncdreamer,
  title={Syncdreamer: Generating multiview-consistent images from a single-view image},
  author={Liu, Yuan and Lin, Cheng and Zeng, Zijiao and Long, Xiaoxiao and Liu, Lingjie and Komura, Taku and Wang, Wenping},
  journal={arXiv preprint arXiv:2309.03453},
  year={2023}
}

@inproceedings{long2024wonder3d,
  title={Wonder3d: Single image to 3d using cross-domain diffusion},
  author={Long, Xiaoxiao and Guo, Yuan-Chen and Lin, Cheng and Liu, Yuan and Dou, Zhiyang and Liu, Lingjie and Ma, Yuexin and Zhang, Song-Hai and Habermann, Marc and Theobalt, Christian and others},
  booktitle={Proceedings of the IEEE/CVF Conference on Computer Vision and Pattern Recognition},
  pages={9970--9980},
  year={2024}
}

@article{hong2023lrm,
  title={Lrm: Large reconstruction model for single image to 3d},
  author={Hong, Yicong and Zhang, Kai and Gu, Jiuxiang and Bi, Sai and Zhou, Yang and Liu, Difan and Liu, Feng and Sunkavalli, Kalyan and Bui, Trung and Tan, Hao},
  journal={arXiv preprint arXiv:2311.04400},
  year={2023}
}

@inproceedings{tang2025lgm,
  title={Lgm: Large multi-view gaussian model for high-resolution 3d content creation},
  author={Tang, Jiaxiang and Chen, Zhaoxi and Chen, Xiaokang and Wang, Tengfei and Zeng, Gang and Liu, Ziwei},
  booktitle={European Conference on Computer Vision},
  pages={1--18},
  year={2025},
  organization={Springer}
}

@inproceedings{huang2024epidiff,
  title={Epidiff: Enhancing multi-view synthesis via localized epipolar-constrained diffusion},
  author={Huang, Zehuan and Wen, Hao and Dong, Junting and Wang, Yaohui and Li, Yangguang and Chen, Xinyuan and Cao, Yan-Pei and Liang, Ding and Qiao, Yu and Dai, Bo and others},
  booktitle={Proceedings of the IEEE/CVF Conference on Computer Vision and Pattern Recognition},
  pages={9784--9794},
  year={2024}
}

@article{zhang2024clay,
  title={CLAY: A Controllable Large-scale Generative Model for Creating High-quality 3D Assets},
  author={Zhang, Longwen and Wang, Ziyu and Zhang, Qixuan and Qiu, Qiwei and Pang, Anqi and Jiang, Haoran and Yang, Wei and Xu, Lan and Yu, Jingyi},
  journal={ACM Transactions on Graphics (TOG)},
  volume={43},
  number={4},
  pages={1--20},
  year={2024},
  publisher={ACM New York, NY, USA}
}

@article{wu2024unique3d,
  title={Unique3d: High-quality and efficient 3d mesh generation from a single image},
  author={Wu, Kailu and Liu, Fangfu and Cai, Zhihan and Yan, Runjie and Wang, Hanyang and Hu, Yating and Duan, Yueqi and Ma, Kaisheng},
  journal={Advances in Neural Information Processing Systems},
  volume={37},
  pages={125116--125141},
  year={2024}
}

@article{li2024craftsman,
  title={CraftsMan: High-fidelity Mesh Generation with 3D Native Generation and Interactive Geometry Refiner},
  author={Li, Weiyu and Liu, Jiarui and Chen, Rui and Liang, Yixun and Chen, Xuelin and Tan, Ping and Long, Xiaoxiao},
  journal={arXiv preprint arXiv:2405.14979},
  year={2024}
}

@article{tang2025geometrictexturalconsistency3d,
  title={Towards Geometric and Textural Consistency 3D Scene Generation via Single Image-guided Model Generation and Layout Optimization},
  author={Tang, Xiang and Li, Ruotong and Fan, Xiaopeng},
  journal={arXiv preprint arXiv:2507.14841},
  year={2025}
}

@misc{huang2025literealitygraphicsready3dscene,
  title={LiteReality: Graphics-Ready 3D Scene Reconstruction from RGB-D Scans},
  author={Zhening Huang and Xiaoyang Wu and Fangcheng Zhong and Hengshuang Zhao and Matthias Nießner and Joan Lasenby},
  year={2025},
  eprint={2507.02861},
  url={https://arxiv.org/abs/2507.02861}
}

@article{liu2025agentic,
  title={Agentic 3D Scene Generation with Spatially Contextualized VLMs},
  author={Liu, Xinhang and Tai, Yu-Wing and Tang, Chi-Keung},
  journal={arXiv preprint arXiv:2505.20129},
  year={2025}
}

@article{achiam2023gpt,
  title={Gpt-4 technical report},
  author={Achiam, Josh and Adler, Steven and Agarwal, Sandhini and Ahmad, Lama and Akkaya, Ilge and Aleman, Florencia Leoni and Almeida, Diogo and Altenschmidt, Janko and Altman, Sam and Anadkat, Shyamal and others},
  journal={arXiv preprint arXiv:2303.08774},
  year={2023}
}

@article{dubey2024llama,
  title={The llama 3 herd of models},
  author={Dubey, Abhimanyu and Jauhri, Abhinav and Pandey, Abhinav and Kadian, Abhishek and Al-Dahle, Ahmad and Letman, Aiesha and Mathur, Akhil and Schelten, Alan and Yang, Amy and Fan, Angela and others},
  journal={arXiv e-prints},
  pages={arXiv--2407},
  year={2024}
}

@article{comanici2025gemini,
  title={Gemini 2.5: Pushing the frontier with advanced reasoning, multimodality, long context, and next generation agentic capabilities},
  author={Comanici, Gheorghe and Bieber, Eric and Schaekermann, Mike and Pasupat, Ice and Sachdeva, Noveen and Dhillon, Inderjit and Blistein, Marcel and Ram, Ori and Zhang, Dan and Rosen, Evan and others},
  journal={arXiv preprint arXiv:2507.06261},
  year={2025}
}

@inproceedings{yu2025metascenes,
  title={METASCENES: Towards Automated Replica Creation for Real-world 3D Scans},
  author={Yu, Huangyue and Jia, Baoxiong and Chen, Yixin and Yang, Yandan and Li, Puhao and Su, Rongpeng and Li, Jiaxin and Li, Qing and Liang, Wei and Zhu Song-Chun and Liu, Tengyu and Huang, Siyuan},
  booktitle={Conference on Computer Vision and Pattern Recognition(CVPR)},
  year={2025}
}

@inproceedings{wen2024ouroboros3d,
  title={Ouroboros3d: Image-to-3d generation via 3d-aware recursive diffusion},
  author={Wen, Hao and Huang, Zehuan and Wang, Yaohui and Chen, Xinyuan and Sheng, Lu},
  booktitle={Proceedings of the Computer Vision and Pattern Recognition Conference},
  pages={21631--21641},
  year={2025}
}

@article{xu2024instantmesh,
  title={Instantmesh: Efficient 3d mesh generation from a single image with sparse-view large reconstruction models},
  author={Xu, Jiale and Cheng, Weihao and Gao, Yiming and Wang, Xintao and Gao, Shenghua and Shan, Ying},
  journal={arXiv preprint arXiv:2404.07191},
  year={2024}
}

@inproceedings{voleti2025sv3d,
  title={Sv3d: Novel multi-view synthesis and 3d generation from a single image using latent video diffusion},
  author={Voleti, Vikram and Yao, Chun-Han and Boss, Mark and Letts, Adam and Pankratz, David and Tochilkin, Dmitry and Laforte, Christian and Rombach, Robin and Jampani, Varun},
  booktitle={European Conference on Computer Vision},
  pages={439--457},
  year={2025},
  organization={Springer}
}

@inproceedings{wang2024crm,
  title={Crm: Single image to 3d textured mesh with convolutional reconstruction model},
  author={Wang, Zhengyi and Wang, Yikai and Chen, Yifei and Xiang, Chendong and Chen, Shuo and Yu, Dajiang and Li, Chongxuan and Su, Hang and Zhu, Jun},
  booktitle={European conference on computer vision},
  pages={57--74},
  year={2024},
  organization={Springer}
}

@inproceedings{liu2024one2345++,
  title={One-2-3-45++: Fast single image to 3d objects with consistent multi-view generation and 3d diffusion},
  author={Liu, Minghua and Shi, Ruoxi and Chen, Linghao and Zhang, Zhuoyang and Xu, Chao and Wei, Xinyue and Chen, Hansheng and Zeng, Chong and Gu, Jiayuan and Su, Hao},
  booktitle={Proceedings of the IEEE/CVF Conference on Computer Vision and Pattern Recognition},
  pages={10072--10083},
  year={2024}
}

@article{wu2024direct3d,
  title={Direct3d: Scalable image-to-3d generation via 3d latent diffusion transformer},
  author={Wu, Shuang and Lin, Youtian and Zhang, Feihu and Zeng, Yifei and Xu, Jingxi and Torr, Philip and Cao, Xun and Yao, Yao},
  journal={Advances in Neural Information Processing Systems},
  volume={37},
  pages={121859--121881},
  year={2024}
}

@article{zhao2024michelangelo,
  title={Michelangelo: Conditional 3d shape generation based on shape-image-text aligned latent representation},
  author={Zhao, Zibo and Liu, Wen and Chen, Xin and Zeng, Xianfang and Wang, Rui and Cheng, Pei and Fu, Bin and Chen, Tao and Yu, Gang and Gao, Shenghua},
  journal={Advances in Neural Information Processing Systems},
  volume={36},
  year={2024}
}

@inproceedings{roessle2024l3dg,
  title={L3dg: Latent 3d gaussian diffusion},
  author={Roessle, Barbara and M{\"u}ller, Norman and Porzi, Lorenzo and Rota Bul{\`o}, Samuel and Kontschieder, Peter and Dai, Angela and Nie{\ss}ner, Matthias},
  booktitle={SIGGRAPH Asia 2024 Conference Papers},
  pages={1--11},
  year={2024}
}

@article{wu2024blockfusion,
  title={Blockfusion: Expandable 3d scene generation using latent tri-plane extrapolation},
  author={Wu, Zhennan and Li, Yang and Yan, Han and Shang, Taizhang and Sun, Weixuan and Wang, Senbo and Cui, Ruikai and Liu, Weizhe and Sato, Hiroyuki and Li, Hongdong and others},
  journal={ACM Transactions on Graphics (TOG)},
  volume={43},
  number={4},
  pages={1--17},
  year={2024},
  publisher={ACM New York, NY, USA}
}

@inproceedings{meng2024lt3sd,
  title={Lt3sd: Latent trees for 3d scene diffusion},
  author={Meng, Quan and Li, Lei and Nie{\ss}ner, Matthias and Dai, Angela},
  booktitle={Proceedings of the Computer Vision and Pattern Recognition Conference},
  pages={650--660},
  year={2025}
}

@inproceedings{liu2024part123,
  title={Part123: part-aware 3d reconstruction from a single-view image},
  author={Liu, Anran and Lin, Cheng and Liu, Yuan and Long, Xiaoxiao and Dou, Zhiyang and Guo, Hao-Xiang and Luo, Ping and Wang, Wenping},
  booktitle={ACM SIGGRAPH 2024 Conference Papers},
  pages={1--12},
  year={2024}
}

@inproceedings{dong2025tela,
  title={Tela: Text to layer-wise 3d clothed human generation},
  author={Dong, Junting and Fang, Qi and Huang, Zehuan and Xu, Xudong and Wang, Jingbo and Peng, Sida and Dai, Bo},
  booktitle={European Conference on Computer Vision},
  pages={19--36},
  year={2025},
  organization={Springer}
}

@inproceedings{li2025craftsman3dhighfidelitymeshgeneration,
  title={Craftsman3d: High-fidelity mesh generation with 3d native diffusion and interactive geometry refiner},
  author={Li, Weiyu and Liu, Jiarui and Yan, Hongyu and Chen, Rui and Liang, Yixun and Chen, Xuelin and Tan, Ping and Long, Xiaoxiao},
  booktitle={Proceedings of the Computer Vision and Pattern Recognition Conference},
  pages={5307--5317},
  year={2025}
}

@article{chen2024meshxlneuralcoordinatefield,
  title={Meshxl: Neural coordinate field for generative 3d foundation models},
  author={Chen, Sijin and Chen, Xin and Pang, Anqi and Zeng, Xianfang and Cheng, Wei and Fu, Yijun and Yin, Fukun and Wang, Zhibin and Yu, Jingyi and Yu, Gang and others},
  journal={Advances in Neural Information Processing Systems},
  volume={37},
  pages={97141--97166},
  year={2024}
}

@misc{chen2024meshanythingartistcreatedmeshgeneration,
      title={MeshAnything: Artist-Created Mesh Generation with Autoregressive Transformers}, 
      author={Yiwen Chen and Tong He and Di Huang and Weicai Ye and Sijin Chen and Jiaxiang Tang and Xin Chen and Zhongang Cai and Lei Yang and Gang Yu and Guosheng Lin and Chi Zhang},
      year={2024},
      eprint={2406.10163},
      archivePrefix={arXiv},
      primaryClass={cs.CV},
      url={https://arxiv.org/abs/2406.10163}, 
}

@misc{wang2024llamameshunifying3dmesh,
      title={LLaMA-Mesh: Unifying 3D Mesh Generation with Language Models}, 
      author={Zhengyi Wang and Jonathan Lorraine and Yikai Wang and Hang Su and Jun Zhu and Sanja Fidler and Xiaohui Zeng},
      year={2024},
      eprint={2411.09595},
      archivePrefix={arXiv},
      primaryClass={cs.LG},
      url={https://arxiv.org/abs/2411.09595}, 
}

@misc{hao2024meshtronhighfidelityartistlike3d,
      title={Meshtron: High-Fidelity, Artist-Like 3D Mesh Generation at Scale}, 
      author={Zekun Hao and David W. Romero and Tsung-Yi Lin and Ming-Yu Liu},
      year={2024},
      eprint={2412.09548},
      archivePrefix={arXiv},
      primaryClass={cs.GR},
      url={https://arxiv.org/abs/2412.09548}, 
}

@inproceedings{he2024neurallightrigunlockingaccurate,
  title={Neural lightrig: Unlocking accurate object normal and material estimation with multi-light diffusion},
  author={He, Zexin and Wang, Tengfei and Huang, Xin and Pan, Xingang and Liu, Ziwei},
  booktitle={Proceedings of the Computer Vision and Pattern Recognition Conference},
  pages={26514--26524},
  year={2025}
}

@inproceedings{gao2025meshartgeneratingarticulatedmeshes,
  title={Meshart: Generating articulated meshes with structure-guided transformers},
  author={Gao, Daoyi and Siddiqui, Yawar and Li, Lei and Dai, Angela},
  booktitle={Proceedings of the Computer Vision and Pattern Recognition Conference},
  pages={618--627},
  year={2025}
}

@article{li2025triposghighfidelity3dshape,
  title={Triposg: High-fidelity 3d shape synthesis using large-scale rectified flow models},
  author={Li, Yangguang and Zou, Zi-Xin and Liu, Zexiang and Wang, Dehu and Liang, Yuan and Yu, Zhipeng and Liu, Xingchao and Guo, Yuan-Chen and Liang, Ding and Ouyang, Wanli and others},
  journal={IEEE Transactions on Pattern Analysis and Machine Intelligence},
  year={2025},
  publisher={IEEE}
}

@inproceedings{ge2022long,
  title={Long video generation with time-agnostic vqgan and time-sensitive transformer},
  author={Ge, Songwei and Hayes, Thomas and Yang, Harry and Yin, Xi and Pang, Guan and Jacobs, David and Huang, Jia-Bin and Parikh, Devi},
  booktitle=ECCV,
  year={2022}
}

@inproceedings{hong2022cogvideo,
  title={Cogvideo: Large-scale pretraining for text-to-video generation via transformers},
  author={Hong, Wenyi and Ding, Ming and Zheng, Wendi and Liu, Xinghan and Tang, Jie},
  booktitle=ICLR,
  year={2023}
}

@inproceedings{kondratyuk2024videopoet,
  title={VideoPoet: A Large Language Model for Zero-Shot Video Generation},
  author={Kondratyuk, Dan and Yu, Lijun and Gu, Xiuye and Lezama, Jose and Huang, Jonathan and Schindler, Grant and Hornung, Rachel and Birodkar, Vighnesh and Yan, Jimmy and Chiu, Ming-Chang and others},
  booktitle={ICML},
  year={2024}
}

@inproceedings{gao2025ca2,
  title={Ca2-VDM: Efficient Autoregressive Video Diffusion Model with Causal Generation and Cache Sharing},
  author={Gao, Kaifeng and Shi, Jiaxin and Zhang, Hanwang and Wang, Chunping and Xiao, Jun and Chen, Long},
  booktitle={ICML},
  year={2025},
  organization={PMLR}
}

@article{hu2024acdit,
  title={ACDiT: Interpolating Autoregressive Conditional Modeling and Diffusion Transformer},
  author={Hu, Jinyi and Hu, Shengding and Song, Yuxuan and Huang, Yufei and Wang, Mingxuan and Zhou, Hao and Liu, Zhiyuan and Ma, Wei-Ying and Sun, Maosong},
  journal={arXiv preprint arXiv:2412.07720},
  year={2024}
}

@inproceedings{jin2024pyramidal,
  title={Pyramidal Flow Matching for Efficient Video Generative Modeling},
  author={Jin, Yang and Sun, Zhicheng and Li, Ningyuan and Xu, Kun and Jiang, Hao and Zhuang, Nan and Huang, Quzhe and Song, Yang and Mu, Yadong and Lin, Zhouchen},
  booktitle=ICLR,
  year={2025}
}

@inproceedings{zhao2025deepmeshautoregressiveartistmeshcreation,
  title={Deepmesh: Auto-regressive artist-mesh creation with reinforcement learning},
  author={Zhao, Ruowen and Ye, Junliang and Wang, Zhengyi and Liu, Guangce and Chen, Yiwen and Wang, Yikai and Zhu, Jun},
  booktitle={Proceedings of the IEEE/CVF International Conference on Computer Vision},
  pages={10612--10623},
  year={2025}
}

@inproceedings{wei2025octgptoctreebasedmultiscaleautoregressive,
  title={Octgpt: Octree-based multiscale autoregressive models for 3d shape generation},
  author={Wei, Si-Tong and Wang, Rui-Huan and Zhou, Chuan-Zhi and Chen, Baoquan and Wang, Peng-Shuai},
  booktitle={Proceedings of the Special Interest Group on Computer Graphics and Interactive Techniques Conference Conference Papers},
  pages={1--11},
  year={2025}
}

@misc{li2025step1x3dhighfidelitycontrollablegeneration,
      title={Step1X-3D: Towards High-Fidelity and Controllable Generation of Textured 3D Assets}, 
      author={Weiyu Li and Xuanyang Zhang and Zheng Sun and Di Qi and Hao Li and Wei Cheng and Weiwei Cai and Shihao Wu and Jiarui Liu and Zihao Wang and Xiao Chen and Feipeng Tian and Jianxiong Pan and Zeming Li and Gang Yu and Xiangyu Zhang and Daxin Jiang and Ping Tan},
      year={2025},
      eprint={2505.07747},
      archivePrefix={arXiv},
      primaryClass={cs.CV},
      url={https://arxiv.org/abs/2505.07747}, 
}

@misc{wu2025direct3ds2gigascale3dgeneration,
      title={Direct3D-S2: Gigascale 3D Generation Made Easy with Spatial Sparse Attention}, 
      author={Shuang Wu and Youtian Lin and Feihu Zhang and Yifei Zeng and Yikang Yang and Yajie Bao and Jiachen Qian and Siyu Zhu and Xun Cao and Philip Torr and Yao Yao},
      year={2025},
      eprint={2505.17412},
      archivePrefix={arXiv},
      primaryClass={cs.CV},
      url={https://arxiv.org/abs/2505.17412}, 
}

@misc{wu2025dipodualstateimagescontrolled,
      title={DIPO: Dual-State Images Controlled Articulated Object Generation Powered by Diverse Data}, 
      author={Ruiqi Wu and Xinjie Wang and Liu Liu and Chunle Guo and Jiaxiong Qiu and Chongyi Li and Lichao Huang and Zhizhong Su and Ming-Ming Cheng},
      year={2025},
      eprint={2505.20460},
      archivePrefix={arXiv},
      primaryClass={cs.CV},
      url={https://arxiv.org/abs/2505.20460}, 
}

@misc{ye2025shapellmomninativemultimodalllm,
      title={ShapeLLM-Omni: A Native Multimodal LLM for 3D Generation and Understanding}, 
      author={Junliang Ye and Zhengyi Wang and Ruowen Zhao and Shenghao Xie and Jun Zhu},
      year={2025},
      eprint={2506.01853},
      archivePrefix={arXiv},
      primaryClass={cs.CV},
      url={https://arxiv.org/abs/2506.01853}, 
}

@misc{lin2025partcrafterstructured3dmesh,
      title={PartCrafter: Structured 3D Mesh Generation via Compositional Latent Diffusion Transformers}, 
      author={Yuchen Lin and Chenguo Lin and Panwang Pan and Honglei Yan and Yiqiang Feng and Yadong Mu and Katerina Fragkiadaki},
      year={2025},
      eprint={2506.05573},
      archivePrefix={arXiv},
      primaryClass={cs.CV},
      url={https://arxiv.org/abs/2506.05573}, 
}

@misc{tang2025efficientpartlevel3dobject,
      title={Efficient Part-level 3D Object Generation via Dual Volume Packing}, 
      author={Jiaxiang Tang and Ruijie Lu and Zhaoshuo Li and Zekun Hao and Xuan Li and Fangyin Wei and Shuran Song and Gang Zeng and Ming-Yu Liu and Tsung-Yi Lin},
      year={2025},
      eprint={2506.09980},
      archivePrefix={arXiv},
      primaryClass={cs.CV},
      url={https://arxiv.org/abs/2506.09980}, 
}

@inproceedings{zhao2025assemblerscalable3dassembly,
  title={Assembler: Scalable 3D Part Assembly via Anchor Point Diffusion},
  author={Zhao, Wang and Cao, Yan-Pei and Xu, Jiale and Dong, Yuejiang and Shan, Ying},
  booktitle={Proceedings of the SIGGRAPH Asia 2025 Conference Papers},
  pages={1--11},
  year={2025}
}

@article{chen2024diffusion,
  title={Diffusion forcing: Next-token prediction meets full-sequence diffusion},
  author={Chen, Boyuan and Mart{\'\i} Mons{\'o}, Diego and Du, Yilun and Simchowitz, Max and Tedrake, Russ and Sitzmann, Vincent},
  journal={Advances in Neural Information Processing Systems},
  volume={37},
  pages={24081--24125},
  year={2024}
}

@article{huang2025self,
  title={Self Forcing: Bridging the Train-Test Gap in Autoregressive Video Diffusion},
  author={Huang, Xun and Li, Zhengqi and He, Guande and Zhou, Mingyuan and Shechtman, Eli},
  journal={arXiv preprint arXiv:2506.08009},
  year={2025}
}

@inproceedings{yang2025longlive,
    title={LongLive: Real-time Interactive Long Video Generation},
    author={Shuai Yang and Wei Huang and Ruihang Chu and Yicheng Xiao and Yuyang Zhao and Xianbang Wang and Muyang Li and Enze Xie and Yingcong Chen and Yao Lu and Song Hanand Yukang Chen},
    year={2025},
    booktitle={arxiv},
}

@misc{liu2025rollingforcingautoregressivelong,
      title={Rolling Forcing: Autoregressive Long Video Diffusion in Real Time}, 
      author={Kunhao Liu and Wenbo Hu and Jiale Xu and Ying Shan and Shijian Lu},
      year={2025},
      eprint={2509.25161},
      archivePrefix={arXiv},
      primaryClass={cs.CV},
      url={https://arxiv.org/abs/2509.25161}, 
}

@article{huang2025memory,
  title={Memory Forcing: Spatio-Temporal Memory for Consistent Scene Generation on Minecraft},
  author={Huang, Junchao and Hu, Xinting and Han, Boyao and Shi, Shaoshuai and Tian, Zhuotao and He, Tianyu and Jiang, Li},
  journal={arXiv preprint arXiv:2510.03198},
  year={2025}
}

@article{cui2025self,
  title={Self-Forcing++: Towards Minute-Scale High-Quality Video Generation},
  author={Cui, Justin and Wu, Jie and Li, Ming and Yang, Tao and Li, Xiaojie and Wang, Rui and Bai, Andrew and Ban, Yuanhao and Hsieh, Cho-Jui},
  journal={arXiv preprint arXiv:2510.02283},
  year={2025}
}

@article{chen2025skyreels,
  title={SkyReels-V2: Infinite-length Film Generative Model},
  author={Chen, Guibin and Lin, Dixuan and Yang, Jiangping and Lin, Chunze and Zhu, Juncheng and Fan, Mingyuan and Zhang, Hao and Chen, Sheng and Chen, Zheng and Ma, Chengchen and others},
  journal={arXiv preprint arXiv:2504.13074},
  year={2025}
}

@article{gu2025long,
  title={Long-Context Autoregressive Video Modeling with Next-Frame Prediction},
  author={Gu, Yuchao and Mao, Weijia and Shou, Mike Zheng},
  journal={arXiv preprint arXiv:2503.19325},
  year={2025}
}

@inproceedings{yin2024one,
  title={One-step diffusion with distribution matching distillation},
  author={Yin, Tianwei and Gharbi, Micha{\"e}l and Zhang, Richard and Shechtman, Eli and Durand, Fredo and Freeman, William T and Park, Taesung},
  booktitle=CVPR,
  year={2024}
}

@article{yin2024improved,
  title={Improved distribution matching distillation for fast image synthesis},
  author={Yin, Tianwei and Gharbi, Micha{\"e}l and Park, Taesung and Zhang, Richard and Shechtman, Eli and Durand, Fredo and Freeman, Bill},
  journal={NeurIPS},
  year={2024}
}

@misc{magi1,
      title={MAGI-1: Autoregressive Video Generation at Scale},
      author={Sand-AI},
      year={2025},
      url={https://static.magi.world/static/files/MAGI_1.pdf},
}

@inproceedings{yin2025causvid,
  title={From Slow Bidirectional to Fast Autoregressive Video Diffusion Models},
  author={Yin, Tianwei and Zhang, Qiang and Zhang, Richard and Freeman, William T and Durand, Fredo and Shechtman, Eli and Huang, Xun},
  booktitle=CVPR,
  year={2025}
}

@inproceedings{qu2025deocc1to33ddeocclusionsingle,
  title={Deocc-1-to-3: 3d de-occlusion from a single image via self-supervised multi-view diffusion},
  author={Qu, Yansong and Dai, Shaohui and Li, Xinyang and Wang, Yuze and Shen, You and Zhang, Shengchuan and Cao, Liujuan},
  booktitle={Proceedings of the AAAI Conference on Artificial Intelligence},
  volume={40},
  number={11},
  pages={8677--8685},
  year={2026}
}

@inproceedings{dong2025morecontextuallatents3d,
  title={From one to more: Contextual part latents for 3d generation},
  author={Dong, Shaocong and Ding, Lihe and Chen, Xiao and Li, Yaokun and Wang, Yuxin and Wang, Yucheng and Wang, Qi and Kim, Jaehyeok and Gao, Chenjian and Huang, Zhanpeng and others},
  booktitle={Proceedings of the IEEE/CVF International Conference on Computer Vision},
  pages={8230--8240},
  year={2025}
}

@inproceedings{huang2025stereogsmultiviewstereovision,
  title={Stereo-GS: Multi-View Stereo Vision Model for Generalizable 3D Gaussian Splatting Reconstruction},
  author={Huang, Xiufeng and Cheung, Ka Chun and Cong, Runmin and See, Simon and Wan, Renjie},
  booktitle={Proceedings of the 33rd ACM International Conference on Multimedia},
  pages={9822--9831},
  year={2025}
}

@misc{chen2025ultra3defficienthighfidelity3d,
      title={Ultra3D: Efficient and High-Fidelity 3D Generation with Part Attention}, 
      author={Yiwen Chen and Zhihao Li and Yikai Wang and Hu Zhang and Qin Li and Chi Zhang and Guosheng Lin},
      year={2025},
      eprint={2507.17745},
      archivePrefix={arXiv},
      primaryClass={cs.CV},
      url={https://arxiv.org/abs/2507.17745}, 
}

@article{ho2020ddpm,
  title={Denoising diffusion probabilistic models},
  author={Ho, Jonathan and Jain, Ajay and Abbeel, Pieter},
  journal={Advances in neural information processing systems},
  volume={33},
  pages={6840--6851},
  year={2020}
}

@article{song2020ddim,
  title={Denoising diffusion implicit models},
  author={Song, Jiaming and Meng, Chenlin and Ermon, Stefano},
  journal={arXiv preprint arXiv:2010.02502},
  year={2020}
}

@inproceedings{deitke2023objaverse,
  title={Objaverse: A universe of annotated 3d objects},
  author={Deitke, Matt and Schwenk, Dustin and Salvador, Jordi and Weihs, Luca and Michel, Oscar and VanderBilt, Eli and Schmidt, Ludwig and Ehsani, Kiana and Kembhavi, Aniruddha and Farhadi, Ali},
  booktitle={Proceedings of the IEEE/CVF Conference on Computer Vision and Pattern Recognition},
  pages={13142--13153},
  year={2023}
}

@article{deitke2024objaversexl,
  title={Objaverse-xl: A universe of 10m+ 3d objects},
  author={Deitke, Matt and Liu, Ruoshi and Wallingford, Matthew and Ngo, Huong and Michel, Oscar and Kusupati, Aditya and Fan, Alan and Laforte, Christian and Voleti, Vikram and Gadre, Samir Yitzhak and others},
  journal={Advances in Neural Information Processing Systems},
  volume={36},
  year={2024}
}

@article{vae,
  title={Auto-encoding variational bayes},
  author={Kingma, Diederik P},
  journal={arXiv preprint arXiv:1312.6114},
  year={2013}
}

@inproceedings{dit,
  title={Scalable diffusion models with transformers},
  author={Peebles, William and Xie, Saining},
  booktitle={Proceedings of the IEEE/CVF International Conference on Computer Vision},
  pages={4195--4205},
  year={2023}
}

@article{li2025triposg,
  title={Triposg: High-fidelity 3d shape synthesis using large-scale rectified flow models},
  author={Li, Yangguang and Zou, Zi-Xin and Liu, Zexiang and Wang, Dehu and Liang, Yuan and Yu, Zhipeng and Liu, Xingchao and Guo, Yuan-Chen and Liang, Ding and Ouyang, Wanli and others},
  journal={IEEE Transactions on Pattern Analysis and Machine Intelligence},
  year={2025},
  publisher={IEEE}
}

@inproceedings{huang2024mvadapter,
  title={Mv-adapter: Multi-view consistent image generation made easy},
  author={Huang, Zehuan and Guo, Yuan-Chen and Wang, Haoran and Yi, Ran and Ma, Lizhuang and Cao, Yan-Pei and Sheng, Lu},
  booktitle={Proceedings of the IEEE/CVF International Conference on Computer Vision},
  pages={16377--16387},
  year={2025}
}

@article{atiss,
  title={Atiss: Autoregressive transformers for indoor scene synthesis},
  author={Paschalidou, Despoina and Kar, Amlan and Shugrina, Maria and Kreis, Karsten and Geiger, Andreas and Fidler, Sanja},
  journal={Advances in neural information processing systems},
  volume={34},
  pages={12013--12026},
  year={2021}
}

@article{layoutgpt,
  title={Layoutgpt: Compositional visual planning and generation with large language models},
  author={Feng, Weixi and Zhu, Wanrong and Fu, Tsu-jui and Jampani, Varun and Akula, Arjun and He, Xuehai and Basu, Sugato and Wang, Xin Eric and Wang, William Yang},
  journal={Advances in Neural Information Processing Systems},
  volume={36},
  pages={18225--18250},
  year={2023}
}

@inproceedings{holodeck,
  title={Holodeck: Language guided generation of 3d embodied ai environments},
  author={Yang, Yue and Sun, Fan-Yun and Weihs, Luca and VanderBilt, Eli and Herrasti, Alvaro and Han, Winson and Wu, Jiajun and Haber, Nick and Krishna, Ranjay and Liu, Lingjie and others},
  booktitle={Proceedings of the IEEE/CVF Conference on Computer Vision and Pattern Recognition},
  pages={16227--16237},
  year={2024}
}

@inproceedings{idesign,
   title={I-Design: Personalized LLM Interior Designer},
   booktitle={Computer Vision – ECCV 2024 Workshops},
   author={Çelen, Ata and Han, Guo and Schindler, Konrad and Van Gool, Luc and Armeni, Iro and Obukhov, Anton and Wang, Xi},
   year={2025},
}

@inproceedings{layoutvlm,
  title={Layoutvlm: Differentiable optimization of 3d layout via vision-language models},
  author={Sun, Fan-Yun and Liu, Weiyu and Gu, Siyi and Lim, Dylan and Bhat, Goutam and Tombari, Federico and Li, Manling and Haber, Nick and Wu, Jiajun},
  booktitle={Proceedings of the Computer Vision and Pattern Recognition Conference},
  pages={29469--29478},
  year={2025}
}

@inproceedings{lin2024instructscene,
  title={InstructScene: Instruction-Driven 3D Indoor Scene Synthesis with Semantic Graph Prior},
  author={Lin, Chenguo and Mu, Yadong},
  booktitle={International Conference on Learning Representations (ICLR)},
  year={2024}
}

@inproceedings{yang2024physcene,
  title={PhyScene: Physically Interactable 3D Scene Synthesis for Embodied AI},
  author={Yang, Yandan and Jia, Baoxiong and Zhi, Peiyuan and Huang, Siyuan},
  booktitle={Proceedings of Conference on Computer Vision and Pattern Recognition (CVPR)},
  year={2024}
}

@article{ling2024scenethesis,
  title         = {Scenethesis: Combining Language and Visual Priors for 3D Scene Generation},
  author        = {Ling, Lu and Lin, Chen-Hsuan and Lin, Tsung-Yi and Ding, Yifan and Zeng, Yu and Sheng, Yichen and Ge, Yunhao and Liu, Ming-Yu and Bera, Aniket and Li, Zhaoshuo},
  journal       = {arXiv preprint arXiv:2505.02836}, 
  year          = {2025}
}

@article{Gu2025ArtiSceneLA,
  title={ArtiScene: Language-Driven Artistic 3D Scene Generation Through Image Intermediary},
  author={Zeqi Gu and Yin Cui and Zhaoshuo Li and Fangyin Wei and Yunhao Ge and Jinwei Gu and Ming-Yu Liu and Abe Davis and Yifan Ding},
  journal={2025 IEEE/CVF Conference on Computer Vision and Pattern Recognition (CVPR)},
  year={2025},
  pages={2891-2901},
  url={https://api.semanticscholar.org/CorpusID:279075256}
}

@inproceedings{SpatialLM,
    title     = {SpatialLM: Training Large Language Models for Structured Indoor Modeling},
    author    = {Mao, Yongsen and Zhong, Junhao and Fang, Chuan and Zheng, Jia and Tang, Rui and Zhu, Hao and Tan, Ping and Zhou, Zihan},
    booktitle = {Advances in Neural Information Processing Systems},
    year      = {2025}
}

@article{Yang2025LLMdrivenIS,
  title={LLM-driven Indoor Scene Layout Generation via Scaled Human-aligned Data Synthesis and Multi-Stage Preference Optimization},
  author={Yixuan Yang and Zhen Luo and Tongsheng Ding and Junru Lu and Mingqi Gao and Jinyu Yang and Victor Sanchez and Feng Zheng},
  journal={ArXiv},
  year={2025},
  volume={abs/2506.07570},
  url={https://api.semanticscholar.org/CorpusID:279251590}
}

@inproceedings{midi,
  title={Midi: Multi-instance diffusion for single image to 3d scene generation},
  author={Huang, Zehuan and Guo, Yuan-Chen and An, Xingqiao and Yang, Yunhan and Li, Yangguang and Zou, Zi-Xin and Liang, Ding and Liu, Xihui and Cao, Yan-Pei and Sheng, Lu},
  booktitle={Proceedings of the Computer Vision and Pattern Recognition Conference},
  year={2025}
}

@inproceedings{scenegen,
    author    = {Meng, Yanxu and Wu, Haoning and Zhang, Ya and Xie, Weidi},
    title     = {SceneGen: Single-Image 3D Scene Generation in One Feedforward Pass},
    booktitle = {arXiv},
    year      = {2025},
}

@article{rope,
  title={Roformer: Enhanced transformer with rotary position embedding},
  author={Su, Jianlin and Ahmed, Murtadha and Lu, Yu and Pan, Shengfeng and Bo, Wen and Liu, Yunfeng},
  journal={Neurocomputing},
  volume={568},
  pages={127063},
  year={2024},
  publisher={Elsevier}
}

@inproceedings{qwen2.5vl,
    title={Qwen2.5-VL Technical Report}, 
    author={Shuai Bai and Keqin Chen and Xuejing Liu and Jialin Wang and Wenbin Ge and Sibo Song and Kai Dang and Peng Wang and Shijie Wang and Jun Tang and Humen Zhong and Yuanzhi Zhu and Mingkun Yang and Zhaohai Li and Jianqiang Wan and Pengfei Wang and Wei Ding and Zheren Fu and Yiheng Xu and Jiabo Ye and Xi Zhang and Tianbao Xie and Zesen Cheng and Hang Zhang and Zhibo Yang and Haiyang Xu and Junyang Lin},
    year={2025},
    booktitle={arXiv},
}

@inproceedings{gpt4o,
    title={GPT-4o System Card}, 
    author={OpenAI teams},
    year={2024},
    booktitle={arXiv},
}

@article{objaversexl,
  title={Objaverse-xl: A universe of 10m+ 3d objects},
  author={Deitke, Matt and Liu, Ruoshi and Wallingford, Matthew and Ngo, Huong and Michel, Oscar and Kusupati, Aditya and Fan, Alan and Laforte, Christian and Voleti, Vikram and Gadre, Samir Yitzhak and others},
  journal={Advances in Neural Information Processing Systems},
  volume={36},
  pages={35799--35813},
  year={2023}
}

@inproceedings{abo,
  title={Abo: Dataset and benchmarks for real-world 3d object understanding},
  author={Collins, Jasmine and Goel, Shubham and Deng, Kenan and Luthra, Achleshwar and Xu, Leon and Gundogdu, Erhan and Zhang, Xi and Vicente, Tomas F Yago and Dideriksen, Thomas and Arora, Himanshu and others},
  booktitle={Proceedings of the IEEE/CVF conference on computer vision and pattern recognition},
  pages={21126--21136},
  year={2022}
}

@article{3dfuture,
  title={3d-future: 3d furniture shape with texture},
  author={Fu, Huan and Jia, Rongfei and Gao, Lin and Gong, Mingming and Zhao, Binqiang and Maybank, Steve and Tao, Dacheng},
  journal={International Journal of Computer Vision},
  volume={129},
  number={12},
  pages={3313--3337},
  year={2021},
  publisher={Springer}
}

@inproceedings{hssd,
  title={Habitat synthetic scenes dataset (hssd-200): An analysis of 3d scene scale and realism tradeoffs for objectgoal navigation},
  author={Khanna, Mukul and Mao, Yongsen and Jiang, Hanxiao and Haresh, Sanjay and Shacklett, Brennan and Batra, Dhruv and Clegg, Alexander and Undersander, Eric and Chang, Angel X and Savva, Manolis},
  booktitle={Proceedings of the IEEE/CVF Conference on Computer Vision and Pattern Recognition},
  pages={16384--16393},
  year={2024}
}

@inproceedings{internscenes,
    title={InternScenes: A Large-scale Simulatable Indoor Scene Dataset with Realistic Layouts}, 
    author={Weipeng Zhong and Peizhou Cao and Yichen Jin and Li Luo and Wenzhe Cai and Jingli Lin and Hanqing Wang and Zhaoyang Lyu and Tai Wang and Bo Dai and Xudong Xu and Jiangmiao Pang},
    year={2025},
    booktitle={arXiv},
}

@inproceedings{qwenimage,
    title={Qwen-Image Technical Report}, 
    author={Chenfei Wu and Jiahao Li and Jingren Zhou and Junyang Lin and Kaiyuan Gao and Kun Yan and Sheng-ming Yin and Shuai Bai and Xiao Xu and Yilei Chen and Yuxiang Chen and Zecheng Tang and Zekai Zhang and Zhengyi Wang and An Yang and Bowen Yu and Chen Cheng and Dayiheng Liu and Deqing Li and Hang Zhang and Hao Meng and Hu Wei and Jingyuan Ni and Kai Chen and Kuan Cao and Liang Peng and Lin Qu and Minggang Wu and Peng Wang and Shuting Yu and Tingkun Wen and Wensen Feng and Xiaoxiao Xu and Yi Wang and Yichang Zhang and Yongqiang Zhu and Yujia Wu and Yuxuan Cai and Zenan Liu},
    year={2025},
    booktitle={arXiv},
}

@inproceedings{mmdit,
  title={Scaling rectified flow transformers for high-resolution image synthesis},
  author={Esser, Patrick and Kulal, Sumith and Blattmann, Andreas and Entezari, Rahim and M{\"u}ller, Jonas and Saini, Harry and Levi, Yam and Lorenz, Dominik and Sauer, Axel and Boesel, Frederic and others},
  booktitle={Forty-first international conference on machine learning},
  year={2024}
}

@inproceedings{cfg,
    title={Classifier-Free Diffusion Guidance}, 
    author={Jonathan Ho and Tim Salimans},
    year={2022},
    booktitle={arXiv},
}

@inproceedings{adamw,
    title={Decoupled Weight Decay Regularization}, 
    author={Ilya Loshchilov and Frank Hutter},
    year={2019},
    booktitle={arXiv},
}

@inproceedings{nash2020polygen,
  title={Polygen: An autoregressive generative model of 3d meshes},
  author={Nash, Charlie and Ganin, Yaroslav and Eslami, SM Ali and Battaglia, Peter},
  booktitle={International conference on machine learning},
  pages={7220--7229},
  year={2020},
  organization={PMLR}
}

@inproceedings{ibing2023octree,
  title={Octree transformer: Autoregressive 3d shape generation on hierarchically structured sequences},
  author={Ibing, Moritz and Kobsik, Gregor and Kobbelt, Leif},
  booktitle={Proceedings of the IEEE/CVF Conference on Computer Vision and Pattern Recognition},
  pages={2698--2707},
  year={2023}
}

@article{li2024pasta,
  title={PASTA: Controllable part-aware shape generation with autoregressive transformers},
  author={Li, Songlin and Paschalidou, Despoina and Guibas, Leonidas},
  journal={arXiv preprint arXiv:2407.13677},
  year={2024}
}

@article{lu2025uni3dar,
  author    = {Shuqi Lu and Haowei Lin and Lin Yao and Zhifeng Gao and Xiaohong Ji and Weinan E and Linfeng Zhang and Guolin Ke},
  title     = {Uni-3DAR: Unified 3D Generation and Understanding via Autoregression on Compressed Spatial Tokens},
  journal   = {Arxiv},
  year      = {2025},
}
}

\end{document}